УДК 519.711.3; 539.1.08; 612.82; 681.322

# НЕЙРОСЕТЕВАЯ АНСАМБЛЕВАЯ МОДЕЛЬ ПАМЯТИ С МАКСИМАЛЬНО ПРАВДОПОДОБНЫМИ ХАРАКТЕРИСТИКАМИ ВСПОМИНАНИЯ И УЗНАВАНИЯ

*Гопыч П.М.*[*]

*Харьковский национальный университет им. В.Н.Каразина, Украина*

Показано, что недавно предложенная нейросетевая модель для определения основных характеристик памяти основана на алгоритме триарно-бинарного кодирования-декодирования, который приводит к новой нейросетевой ансамблевой модели памяти (НСАМП) с максимально-правдоподобными характеристиками вспоминания-узнавания и новой архитектурой ячейки памяти, включающей хопфилдовскую двухслойную сеть, многоканальные временные ворота, вспомогательную эталонную память и две вложенные петли обратной связи. Найдены условия, при которых один из вариантов сети Хопфилда реализует для используемого алгоритма кодирования максимально правдоподобный алгоритм декодирования типа свертки и, одновременно, алгоритм классификации произвольных бинарных векторов по отношению к значению их хемминговского расстояния до заданного эталона. В дополнение к основным характеристикам памяти и т.д. модель явно описывает зависимость от времени процесса извлечения информации из памяти, а также предоставляет возможность одношагового обучения, моделирования метапамяти, обобщенного представления знаний и раздельного описания сознательных и подсознательных ментальных процессов. Показано, что ячейку ансамблевой памяти можно рассматривать как модель наименьшего неделимого элемента, «атома» сознания. Обсуждаются применения НСАМП для решения некоторых междисциплинарных задач из разных областей знания и ряд нетрадиционных нейробиологических аргументов в пользу ее правдоподобности (явление динамической пространственно-временной синхронизации, свойства нейронов зависящих от времени и нейронов-детекторов ошибок, ранняя прецизионно синхронная деполяризация нейронов и т.д.).

It has been shown that a neural network model recently proposed to describe basic memory performance is based on a ternary/binary coding/decoding algorithm which leads to a new neural network assembly memory model (NNAMM) providing maximum-likelihood recall/recognition properties and implying a new memory unit architecture with Hopfield two-layer network, *N*-channel time gate, auxiliary reference memory, and two nested feedback loops. For the data coding used, conditions are found under which a version of Hopfied network implements maximum-likelihood convolutional decoding algorithm and, simultaneously, linear statistical classifier of arbitrary binary vectors with respect to Hamming distance between vector analyzed and reference vector given. In addition to basic memory performance and etc, the model explicitly describes the dependence on time of memory trace retrieval, gives a possibility of one-trial learning, metamemory simulation, generalized knowledge representation, and distinct description of conscious and unconscious mental processes. It has been shown that an assembly memory unit may be viewed as a model of a smallest inseparable part or an 'atom' of consciousness. Some nontraditional neurobiological backgrounds (dynamic spatiotemporal synchrony, properties of time dependent and error detector neurons, early precise spike firing, etc) and the model's application to solve some interdisciplinary problems from different scientific fields are discussed.

## ВВЕДЕНИЕ

В работе [1], используя искусственные нейронные сети (ИНС) типа Хопфилда [2] и наши результаты [3,4] относящиеся к компьютерным методам автоматического анализа линейчатых спектров излучения, была предложена модель для вычисления основных характеристик памяти в зависимости от интенсивности подсказки. В настоящей работе продемонстрировано, что триарно-бинарный алгоритм кодирования-декодирования [1] является *максимально правдоподобным* и приводит фактически к новой *нейросетевой ансамб-*

---

[*] E-mail: pmg@kharkov.com

*левой модели памяти* (НСАМП) с *максимально правдоподобными* характеристиками вспоминания и узнавания. Для используемого метода кодирования найдены условия, при которых вариант хопфилдовских ИНС [2] реализует максимально правдоподобный алгоритм декодирования типа свертки и, одновременно, статистический линейный классификатор произвольных бинарных векторов по отношению к их хемминговскому расстоянию до заданного эталонного вектора. Для предложенной модели в дополнение к активным и пассивным следам памяти, максимально правдоподобным основным характеристикам памяти, способам описания закрепления и ослабления памяти, свободным параметрам модели и численным примерам обсуждаются активизация конкретной ансамблевой памяти и ее одношаговое обучение, зависящий от времени механизм извлечения информации из памяти, связь НСАМП с другими подходами (в частности, с хопфилдовскими сетями, сетями с распределенным кодированием, с ИНС типа свертки и модульными структурно-составными ИНС) и показано, что компьютерная реализация модели непосредственно требует новой архитектуры ячейки памяти, включающей *двухслойную хопфилдовскую ИНС*, *N-канальные временные ворота*, дополнительную (вспомогательную) *эталонную память* и *две вложенные петли обратной связи*. Учтен ряд нетрадиционных нейробиологических аргументов в пользу предложенной модели: явление динамической пространственно-временной синхронизации, размер основной сигнальной ячейки нейронов, распределенные (популяционные) всплески нервных импульсов и их ранняя прецизионная синхронизация, устойчивое распространение групп нервных импульсов, свойства нейронов зависящих от времени и нейронов-детекторов ошибок. Модель допускает возможность интерпретации ячейки НСАМП как наименьшего неделимого элемента сознания, позволяет учесть осознаваемые (экспликативные) и неосознаваемые (импликативные) взаимодействия конкретной памяти с ее внешним окружением и предложить описание некоторых когнитивных функций мозга, которые до сих пор были вне рамок современных компьютерных моделей. Не касаясь очевидных применений модели для максимально правдоподобного кодирования-декодирования данных или для максимально-правдоподобного запоминания-извлечения из памяти, мы рассматриваем применения модели для описания явлений из различных областей знания, включая моделирование метапамяти, обобщенное представление знаний и раздельное описание сознательных и подсознательных ментальных процессов у человека. В частности, обсуждаются некоторые задачи, относящиеся к *распознаванию образов и теории зрения*, *нейропсихологии*, *психолингвистики* и *психологии эмоций и чувств*. Эффективность модели для решения различных междисциплинарных задач рассматривается как одно из доказательств ее правильности.

Настоящая работа является изложением доклада, представленного на V-м Международном конгрессе по математическому моделированию в Дубне [5].

## 1. ТРИАРНО-БИНАРНОЕ КОДИРОВАНИЕ ДАННЫХ

Будем записывать исходные данные в виде векторов, значения компонент которых могут быть из состава триплета −1, 0, 1. Такие триарные векторы или сигнальные пакеты импульсов соответствуют группам действующих одновременно (в пределах заданного временного окна) нервных импульсов или «спайков», а их компоненты моделируют отсутствие (0) или наличие (±1) спайков, воздействующих на возбудительные (+1) или замедлительные (−1) синапсы целевых нейронов (т.е. знак ненулевой проекции вектора определяет дальнейшее функциональное предназначение соответствующего ему конкретного спайка). Полное число компонент такого вектора велико, так как центральная нервная система содержит очень много нейронов, но большинство из них имеет значение равное нулю, так как в любой заданный момент времени большинство этих нейронов «дремлют», находясь в состоянии покоя. Эта ситуация соответствует случаю *распределенного* (разбросанного или разреженного) кодирования [6] и мы будем обозначать размерность пространства таких векторов $N_{sps}$. Предполагаем, что спящие нейроны не несут информации, существенной в данный момент времени, и, следовательно, могут быть исключены из текущего рассмотрения. Такое исключение является фактически преобразованием исходных векторов из представления распределенного кодирования в представление плотного кодирования, когда учитываются только информативные (ненулевые) компоненты исходного вектора (рис. 1). Будем предполагать, что преобразование триарных векторов в бинарные

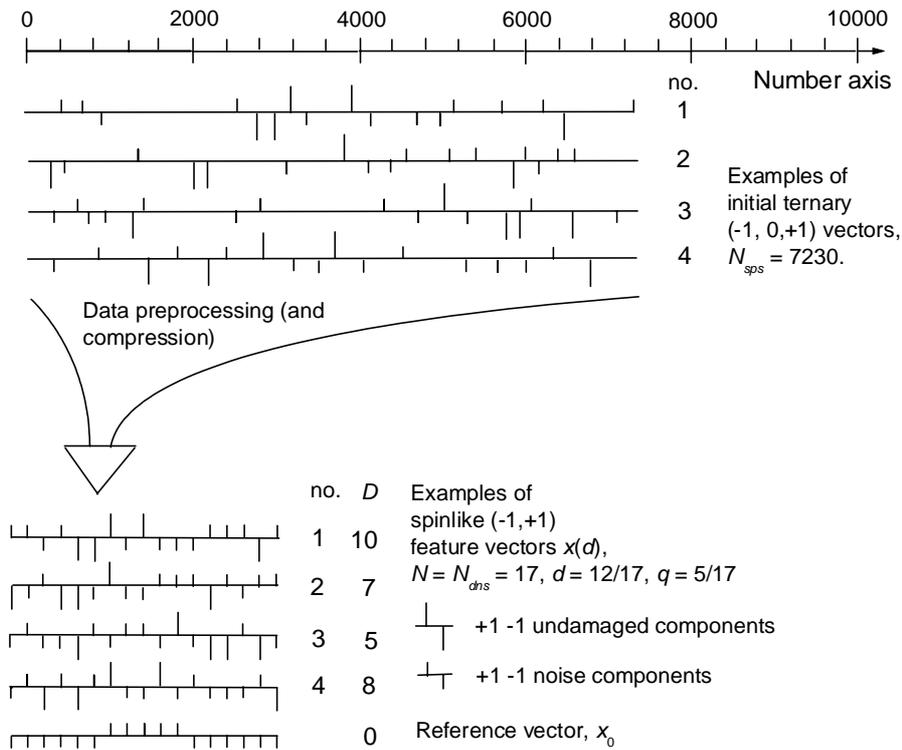

**Рис. 1.** Схема преобразования исходных триарных векторов в квазибинарные характеристические векторы $x(d)$. Показаны четыре примера и их параметры. $D$ обозначает хемминговское расстояние между конкретным вектором $x(d)$ и эталоном $x_0$, $d$ – степень искажения $x_0$, $q$ – интенсивность подсказки, $N_{sps}$ и $N_{dns}$ – размерности разбросанно (распределенно) кодированных исходных векторов и плотно кодированных характеристических векторов, сответственно. Для всех векторов значения $N_{sps}$, $N_{dns}$, $d$ и $q$ одинаковы; их компоненты ±1 показаны как вертикальные черточки, размещенные выше (+) или ниже (–) горизонтальных линий. Более длинные черточки обозначают неискаженные проекции эталона $x_0$, более короткие – компоненты шума. Нулевым компонентам триарных векторов соответствуют точки (по $N_{sps} - N_{dns} = 7213$ на вектор), которые на рисунке сливаются в сплошную линию.

происходит на этапе их *предобработки* (разд. 4.4 и 13.2), когда отбираются только те спайки из потока исходных данных, которые попадают во временное окно совпадений конкретного клеточного ансамбля, выделяемого механизмом динамической пространственно-временной синхронизации (разд. 4.4, 5, 12.4 и 13.1). Размерность $N_{dns}$ плотно кодированного бинарного *характеристического вектора*, получаемого в результате предобработки исходного распределенно кодированного вектора, много меньше чем $N_{sps}$ (рис. 1, разд. 4.1, 5 и 12.3). Причем характеристический вектор является фактически не бинарным, а квазибинарным так как, во-первых, его спиноподобные (±1) компоненты не могут быть преобразованы в другое (0, 1) бинарное представление путем переопределения порогов и констант связи и, во-вторых, все его компоненты имеют третью (нулевую) проекцию, которая здесь не проявляется, но которая существенна, например, при определении наруше ний памяти (разд. 8). Далее будем рассматривать только квазибинарные векторы размерности $N$ ($N = N_{dns}$), но для краткости приставку «квази» в их наименовании будем опускать.

Пусть $x$ это *произвольный вектор* с компонентами $x_i$, $i = 1,...,N$, значения которых могут быть только −1 или +1. Такой вектор может нести $N$ бит информации, а его размерность $N$ – это размер локального *рецептивного поля* (разд. 12.3) или размер (емкость) будущей нейросетевой *ячейки памяти* (разд. 3 и 4). Если $x$ несет ту информацию, которая хранится или должна хранится в нейросетевой ячейке памяти, то будем называть его *эталонным вектором* $x_0$ (как будет видно из разд. 2 и 4.1, для хемминговского статистического классификатора образов $x_0$ является также *оптимальным характеристическим вектором*). Если знаки всех компонент вектора $x$ выбраны случайно и равновероятно, то это *случайный вектор* $x_r$ или бинарный «шум». Введем также *поврежденный* (искаженный или испорченный) *эталонный вектор* $x(d)$ со *степенью повреждения* эталонного вектора $x_0$ равной $d$. Его компоненты $x_i(d)$ определим соотношением

$$x_i(d) = \begin{cases} x_0^i, & \text{if } u_i = 0, \\ x_r^i, & \text{if } u_i = 1 \end{cases} \qquad i = 1,...,N \qquad (1)$$

где $u_i$ – признаки, значения которых 0 или 1 выбираются случайно и равновероятно так, чтобы сохранялось значение степени искажения

$$d = \sum u_i / N, \quad i = 1,...,N. \qquad (2)$$

Ясно, что согласно определенному выше правилу кодирования (правилу учета шума), $x_0$ и $x_r$ неравноправны и шум $x_r$ имеет приоритет над эталоном $x_0$. Если в (1) и (2) число признаков $u_i = 1$ равняется $m$, то $d = m/N$. Кроме того, $0 \le d \le 1$, $x(0) = x_0$, $x(1) = x_r$.

Если вектор $x = x(d)$ содержит часть $q$ неискаженной информации об $x_0$, то $x(d) = x(1 - q)$, где $q$ это *интенсивность* (или *индекс*) *подсказки*, которая определяет полное количество неискаженной информации, которую несет в себе вектор $x(d)$:

$$q = 1 - d. \qquad (3)$$

Ясно, что $q = 1 - m/N$, $0 \le q \le 1$, $x(q) = x(1 - d)$, а $d$ и $q$ дискретны [1].

Для $d = m/N$ число различных векторов $x(d)$ есть $2^m C_m^N$. Если $d$ берется из диапазона значений $0 \le d \le 1$, то набор всех $x(d)$ – это полный *конечный* набор всех возможных разных векторов $x$ размерности $N$ (их количество есть $\sum 2^m C_m^N = 3^N$, где суммирование выполняется по $m = 0,1,...,N$). Поскольку $d = m/N$ есть отношение числа $m$ шумовых компонент вектора $x(d)$ к $N$ (число компонент эталонного вектора $x_0$), то для каждого $x(d)$ $d$ можно интерпретировать как *отношение шум/сигнал*. Поскольку $q = (N - m)/N$ есть отношение числа $N - m$ неискаженных шумом (всегда совпадающих с проекциями эталона $x_0$) компонент вектора $x(d)$ к $N$ (число компонент шумового вектора $x_r$), то для каждого $x(d)$ $q = 1 - d$ можно интерпретировать как *отношение сигнал/шум*. Интенсивность подсказки $q$ есть также степень сходства, относительная средняя (стандартизированная) свертка и *коэффициент корреляции* между векторами $x_0$ и $x(d)$, для которых $N - m$ соответствующих компонент всегда совпадают, а $m$ соответствующих компонент могут различаться по знаку случайно и равновероятно (разд. 2). Отметим также, что в нашем случае отношение сигнал/шум или шум/сигнал нельзя определить как $q/d$ или $d/q$, так как $q/d$ не существует если $d = 0$ (т.е. $q = 1$), а $d/q$ не существует если $q = 0$ (т.е. $d = 1$). Несмотря на это для каждого вектора $x(d)$ сигнал и шум аддитивны в том смысле, что $q + d = 1$, в отличие от обычного случая $q = 1/d$.

Векторы *x*(*d*) не могут иметь нулевых проекций и этим существенно отличаются от обычно используемых векторов состояний ИНС. Действительно, обычно шумовая часть вектора состояния ИНС представляет собой ту часть анализируемых данных, которая испорчена. Ее определяют как случайную последовательность нулей и единиц, если данные кодированы в бинарной форме (0, 1), или как последовательность нулей, если они кодированы в триарной форме (–1, 0, 1) [7]. В отличие от этого, в нашем случае шум есть естественная, внутренне присущая анализируемым данным их неотъемлемая часть (свойство), а не результат замещения испорченных фрагментов данных искусственно приготовленным суррогатом.

Простой способ получения бинарных данных в виде последовательностей –1 и +1 предложен в [4], где рассматривались линейчатые спектры излучения. Там же было установлено, что в ходе бинаризации спектров (и, следовательно, их сжатия) не происходит потери той информации, которая существенна для их декодирования с помощью соответствующего нейросетевого алгоритма (см. разд. 2). Следовательно, имеется широкий класс задач (выделение локальных особенностей из полутоновых изображений, искаженных аддитивным шумом), где естественно возникают обсуждавшиеся выше спиноподобные бинарные данные.

## 2. МАКСИМАЛЬНО ПРАВДОПОДОБНОЕ ДЕКОДИРОВАНИЕ ДАННЫХ

Теперь для предложенного метода кодирования введем соответствующий ему алгоритм декодирования, т.е. правило для извлечения эталонного вектора (образа) $x_0$ из исходного вектора $x_{in} = x(d)$, который может интерпретироваться как шум или как эталон $x_0$ искаженный, со степенью его искажения $d$, фоном $x_r$.

Рассмотрим двухслойную автоассоциативную ИНС с $N$ нейронами во входном (выходном) слое. Все нейроны (синаптические памяти) входного слоя ИНС связаны с нейронами ее выходного слоя по правилу «все со всеми». В качестве синаптических памятей используем простые нейроны МакКаллоха-Питса со ступенчатой ответной функцией, нулевым порогом срабатывания и обобщенным обучающим правилом по Хеббу.

Следуя [2], для обученной ИНС элементы $w_{ij}$ ее *синаптической матрицы* $w$ определим согласно правилу

$$w_{ij} = \eta\, x_0^i x_0^j \tag{4}$$

где $\eta > 0$ – *параметр обучения* (здесь и далее удобно предполагать, что $\eta = 1$), а $x_0^i, x_0^j$ с $i,j = 1,..,N$ – компоненты эталонного вектора $x_0$ (все $w_{ij}$ отличаются друг от друга только знаком). Таким образом, на основе информации, которая должна быть запомнена в памяти (вектор $x_0$), уравнение (4) допускает *однозначное* вычисление $w$. Будем называть сеть с синаптической матрицей $w$ идеально обученной ИНС. Такая сеть помнит *только один* эталонный образ $x_0$ и мы *намеренно* пренебрегаем имеющейся возможностью [2] запоминания других следов памяти в той же ИНС.

Введенная ИНС обслуживается *N*-мерными векторами *x*. Предполагаем, что входной вектор $x_{in}$ декодирован (эталонный вектор $x_0$ извлечен из $x_{in}$) успешно, если обученная ИНС преобразует $x_{in}$ в выходной вектор $x_{out} = x_0$. Алгоритм преобразования следующий.

*Входной сигнал* $h_j$ для *j*-го нейрона выходного слоя ИНС есть

$$h_j = \sum w_{ij} v_i + s_j \tag{5}$$

где $v_i$ есть выходной сигнал *i*-го нейрона входного слоя сети; $s_j = 0$ – сигнал смещения.

*Выходной сигнал* $v_j$ для *j*-го нейрона выходного слоя сети (*j*-я компонента вектора $x_{out}$) вычисляется с помощью сигмоидной функции (однобитового квантизатора)

$$v_j = \begin{cases} +1, & \text{if } h_j > 0 \\ -1, & \text{if } h_j \leq 0 \end{cases} \quad (6)$$

где для $h_j = 0$ значение $v_j = -1$ (а не +1) выбрано произвольно. О некоторых последствиях такого выбора см. разд. 11.

Если $h_i$ есть *i*-я компонента $x_{in}$, то сигмоидная функция определяет $v_i$ для *i*-го нейрона входного слоя сети. Следовательно, для $h_i = x^i_{in}$ и $x_{in}$ с компонентами ±1 имеем $v_i = x^i_{in}$. Из этого факта и уравнений (4) и (5) следует, что для *j*-го выходного нейрона $h_j = \Sigma w_{ij} x^i_{in} = \eta x^j_0 (\Sigma x^i_0 x^i_{in}) = \eta Q x^j_0$, где $\eta$ – обучающий параметр, а $Q = \Sigma x^i_0 x^i_{in}$ – *свертка* векторов $x_0$ и $x_{in}$. Так как для каждого вектора $x_{in}$ существует такой вектор $x(d)$, что $x_{in} = x(d)$, то свертку $Q$ можно записать как $Q(d)$. После подстановки $h_j = \eta Q x^j_0$ в выражение (6) найдем, что $x_{out} = x_0$ и эталонный вектор идентифицирован успешно, если $Q(d)$ больше нуля:

$$Q(d) = \sum x_i(d) x^i_0 > 0. \quad (7)$$

Таким образом, нейросетевой алгоритм декодирования, основанный на уравнениях (5) и (6), эквивалентен более простому и быстрому алгоритму декодирования типа свертки (впервые это было отмечено в [4]). Неравенство (7) отбирает такие векторы $x(d)$, для которых число их компонент, совпадающих с соответствующими проекциями $x_0$, больше $N/2$. Но аналитическое выражение для $Q(d)$, используемое в (7), справедливо только для неповрежденных сетей. Если сеть повреждена, то для каждого распределения ее локальных повреждений (см. разд. 8) соответствующее выражение для $Q(d)$ нужно находить отдельно и, вероятно, это главный недостаток описанного конволюционного или «сверточного» подхода (например, количество $n$ разных вариантов выбора $k$ разрываемых межнейронных связей для сети с $N$ нейронами во входном и выходном слое есть $n = C^{N \times N}_k$ и даже при не очень больших $N$ и $k$ оно может быть огромно).

Выше для модельных нейронов выбирался порог их срабатывания $l = 0$, в то время как $l$ могут быть, вообще говоря, и положительными, и отрицательными из диапазона, определяемого значением $N$. Полное количество разных значений $Q$, удовлетворяющих двойному неравенству $-N \leq Q \leq N$, есть $N + 1$ и разница между любыми двумя соседними значениями $Q$ есть $\Delta Q = 2$. Если предположить, что $l \neq 0$, то в (6) вместо $h_j > 0$ и $h_j \leq 0$ следует писать $h_j > l$ и $h_j \leq l$, соответственно, и тогда вместо (7) получим $Q > l$ (если $\eta \neq 1$, то $Q > l/\eta$). Следовательно, $Q$ и $h_j$ функционально связаны (см. также выше), $Q$ можно рассматривать как некий сигнал и, следовательно, нейробиологический смысл неравенства (7) состоит в сравнении этого сигнала с порогом срабатывания модельных нейронов $l$.

Непосредственно легко получить, что $Q(d) = N - 2D(d)$ и $D(d) = (N - Q(d))/2$, где расстояние Хемминга $D(d)$ между векторами $x_0$ и $x(d)$ определяется как число их соответствующих компонент, которые различаются по знаку. Поскольку между $Q$ и $D$ существует взаимно однозначное соответствие, то вместе с неравенством (7) верно и неравенство $D(d) < N/2$. Это означает, что сверточный алгоритм (7), или нейросетевой алгоритм (5) и (6), *непосредственно* отбирают те образы $x(d)$, которые ближе к $x_0$ чем заданное хемминговское расстояние между ними. Более того, $Q(d)$ можно интерпретировать просто как удобное аналитическое выражение для вычисления $D(d)$. Описанный конволюционный или хемминговский алгоритм классификации получен из хопфилдовского нейросетевого алгоритма (4)-(6) без какой бы то ни было дополнительной оптимизационной процедуры и является точным в том смысле, что для данных, которые кодированы как это описано в разд. 1, невозможно создать другой алгоритм с лучшими характеристиками хемминговской классификации и все вероятностные характеристики такой классификации (декодирова ния) векторов могут быть вычислены *точно* (см. также разд. 7 и 11). Другими словами, (7) есть линейный классификатор Хемминга, который по заложенному в нем критерию разделяет векторы $x(d)$ на два класса: набор по-разному искаженных шумом эталонных векторов $x_0$ и набор разных случайных реализаций чистого бинарного шума.

Почленное деление на $N$ и усреднение соотношения $D = (N - Q)/2$ по всем $x(d)$ дает $<D(d)>/N = (1 - <Q(d)>/N)/2$, где $d = m/N$ а угловые скобки обозначают усреднение. Вспоминая, что $d = 1 - q$, имеем $\delta = (1 - \rho)/2$, где среднее относительное хемминговское расстояние между $x_0$ и разными векторами $x(d)$ есть $\delta(d) = <D(d)>/N = d/2 = (1 - q)/2$, а корреляционный коэффициент между ними есть $\rho(d) = <Q(d)>/N = 1 - d = q$ (см. также разд. 1). Таким образом, для вектора $x_0$ и множества векторов $x(d)$ ожидаемое число и ожидаемая доля их соответствующих компонент, которые отличаются, равны $<D> = m/2$ и $\delta = d/2$, а ожидаемое число и ожидаемая доля их соответствующих компонент, которые всегда совпадают, равны $<Q> = N - m$ и $\rho = q$, соответственно. Здесь принято во внимание, что знаки $N - m$ соответствующих компонент $x_0$ и $x(d)$ всегда совпадают, а знаки $m$ оставшихся компонент $x(d)$ выбираются случайно и независимо с вероятностью ½.

Максимально правдоподобный алгоритм декодирования (распознавания, классификации) должен отбирать случаи, когда вероятность $p(d)$ воспроизведения произвольного входного вектора $x_{in} = x(d)$ эталонным вектором $x_0$ и шумом $x_r$ больше, чем вероятность воспроизведения $x(d)$ только шумом. Другими словами, $p(d)$ это вероятность того, что конкретный вектор $x(d)$ есть вектор $x_0$ искаженный со степенью его искажения $d$ шумом $x_r$: $p(d) = 1/2^{dN}$. Поскольку $p(1) = 1/2^N$ есть вероятность того, что $x(d)$ является образцом чистого шума, то $L(d) = p(d)/p(1)$ есть отношение правдоподобия и для максимально правдоподобного алгоритма декодирования должно быть $L = 2^{(1-d)N} = 2^{qN} > 1$. Неравенство $L > 1$ справедливо при всех $d < 1$ (или $q > 0$) и чем меньше $d$ (чем больше $q$) тем больше $L$. Таким образом, $L(d)$ зависит только от подсказки $q = 1 - d$, которая есть степень подобия между $x_0$ и $x(d)$. Но именно $Q(d)$ (или $D(d)$) является выражением для вычисления меры схожести между $x_0$ и конкретным вектором $x(d)$. Это означает, что правило $Q(d) > l$ реализует упомянутый максимально правдоподобный алгоритм декодирования и теперь надо только определить пороговое значение $Q$ (т.е. $l$). Поскольку значение $l$ определяет, в относительных единицах, уровень значимости максимально правдоподобного алгоритма декодирования, то оно может быть выбрано из своей области значений произвольно. Как и в неравенстве (7), по причине нейробиологической правдоподобности такого выбора, для $l$ удобно принять значение $l = 0$.

Таким образом, для неповрежденных ИНС с синаптической матрицей (4) нейросетевой алгоритм (5) и (6) и конволюционный (хемминговский) алгоритм (7) являются вариантами одного и того же оптимального максимально-правдоподобного алгоритма декодирования. Более того, для локально поврежденных ИНС соответствующие алгоритмы декодирования также могут быть максимально правдоподобными, во всяком случае, если их повреждения не являются катастрофически большими (см. разд. 8). Правда, в таких случаях для каждого конкретного распределения локальных повреждений сети соответствующее аналитическое выражение для вычисления свертки $Q$ (или хемминговского расстояния $D$) должно быть получено отдельно. Это также означает, что нейросетевой алгоритм идентификации *PsNet* [4] тоже реализует максимально правдоподобный алгоритм декодирования. Последнее утверждение находится в согласии с тем фактом, что характеристики качества программ распознавания *Alisa* [3] и *PsNet* [4], найденные эмпирически с помощью модельного эксперимента, практически совпадают [4] (*Alisa* реализует традиционную форму максимально правдоподобного алгоритма распознавания типа свертки [3]).

Итак, описанные нейросетевой, конволюционный и хемминговский алгоритмы декодирования (распознавания или классификации) эквивалентны и максимально-правдоподобны.

## 3. ОСНОВЫ НЕЙРОСЕТЕВОЙ АНСАМБЛЕВОЙ МОДЕЛИ ПАМЯТИ

Описанный метод бинарного *кодирования-декодирования* можно легко переформулировать как метод *запоминания-извлечения* из памяти или как нейросетевую *модель памяти*. Для этого в разд. 1 и 2 вместо кодирования и декодирования достаточно просто говорить о запоминании и извлечении из памяти, соответственно. Следовательно, основная идея настоящей работы состоит в том, чтобы построить нейросетевую модель памяти из тех простых объектов с известными свойствами, которые были определены в рамках подхода к кодированию-декодированию, описанному выше.

Соответствующая нейросетевая модель памяти строится из однотипных, взаимосвязанных (ассоциированных) и равноправных ансамблевых ячеек памяти и по причинам, описанным в разд. 13.1, будем называть ее *нейросетевой ансамблевой моделью памяти* (НСАМП) или просто *ансамблевой памятью*. Каждая ячейка ансамблевой памяти (разд. 4) содержит в качестве своей составной части обычную хопфилдовскую ИНС (разд. 2), а также другие функционально новые и по-новому организованные элементы, среди которых *N-канальные временные ворота* (разд. 4.4), *эталонная память* (разд. 4.5) и *две вложенные петли обратной связи* (разд. 4.6). *A priori* предполагаем (см. разд. 2), что ИНС из состава ансамблевой ячейки памяти обучена помнить только *один след* памяти $x_0$ и что запомненная информация извлечена из памяти успешно, если входной вектор $x_{in} = x(d)$ ведет к появлению (инициирует всплывание) на выходе ИНС вектора $x_{out} = x_0$. Предполагаем, что для извлечения из памяти (вспоминания) следа $x_0$, ИНС должна тестироваться *серией* отчасти случайных векторов $x(d)$ до тех пор, пока не всплывет выходной вектор $x_{out} = x_0$ (успешное вспоминание) или процесс вспоминания не будет остановлен по независимым внешним причинам (безуспешное вспоминание). В отличие от *одношагового* процесса декодирования, когда анализируется только один заданный набор исходных данных $x_{in}$, модель памяти предполагает, что при извлечении следа памяти используется *серия* случайно генерируемых (при постоянном значении $d$) *разных* входных векторов $x_{in} = x(d)$, каждый из которых инициирует один шаг этого многошагового процесса. Вследствие свойств исходного алгоритма декодирования, предлагаемый *механизм извлечения информации из памяти* и его *характеристики* (показатели качества) являются оптимальными в том смысле, как это описано в разд. 2.

Нейробиологические основы и нейробиологические аналоги НСАМП в целом и ее отдельных составных элементов обсуждаются в разд. 13.

## 4. АРХИТЕКТУРА ЯЧЕЙКИ АНСАМБЛЕВОЙ ПАМЯТИ

Ячейка ансамблевой памяти конструируется (рис. 2) из синхронизованных на короткое время *блоков 1-6*, их внутренних и внешних *путей* для распространения синхронизованных групп сигналов и *связей* для передачи (асинхронной) управляющей информации. Это становится возможным, если предположить, что все элементы ячейки ансамблевой памяти и ее специфическое окружение выделяются посредством механизма динамической пространственно-временной синхронизации (см. разд. 5 и 13.1).

В *блоке 1* предварительной обработки исходных данных (как будет видно из разд. 4.4 его можно рассматривать как *N-канальные временные ворота*) происходит преобразование исходных триарных распределенно кодированных векторов в двоичные спиноподобные плотно кодированные векторы (разд. 1). Здесь из потока, вообще говоря, не синхронизованных входных сигналов (который может содержать группы синхронизованных спайков, поступающих как от первичных сенсорных входов, так и от других ансамблевых ячеек памяти), приготавливается синхронизованный пакет сигналов в виде *N-мерного характеристического вектора* $x_{in}$ (разд. 1, 2 и 4.1). *Блок 2* – это нейросетевая ячейка памяти, обу ченная согласно уравнению (4) из разд. 2 или (10) из разд. 10, в которой с помощью уравнений (5) и (6) каждый входной вектор $x_{in}$ преобразуется в соответствующий ему выходной вектор $x_{out}$. *Блок 3* выполняет сравнение только что всплывшего вектора $x_{out}$ с *эталонным следом памяти $x_0$ из эталонной памяти* (разд. 4.5). Если $x_{out} = x_0$, то процесс вспоминания успешно завершен и на этом прекращается. В противном случае активируется внутренняя *импликативная* петля обратной связи 1-2-3-4-1 (см. разд. 4.6), процесс извлечения из памяти начинается с *блока 1* заново и повторяется пока текущее значение *времени вспоминания t* меньше его заданного максимально возможного значения $t_0$ (это проверяется в *блоке 4*). Если время $t_0$ – характерный параметр нейронов зависящих от времени (разд. 13.5) – оказывается недостаточным для успешного вспоминания (извлечения из памяти) следа $x_0$, то в *блоке 5* (о *блоках 3,4* и *5* см. разд. 4.3) проверяется, существует ли независимая внешняя причина для продолжения вспоминания. Если да, то активируется внешняя *экспликативная* петля обратной связи 1-2-3-4-5-6-1 (разд. 4.6), счет времени начинается заново (*блок 6*) и внутренний цикл 1-2-3-4-1 опять повторяется (с частотой $f$ или периодом $1/f$) пока $t < t_0$. Если частота $f$ фиксирована, то для цикла 1-2-3-4-1 $ft$ и $ft_0$ являются текущим номером вектора $x_{in}$ и его максимально возможным номером, соответственно.

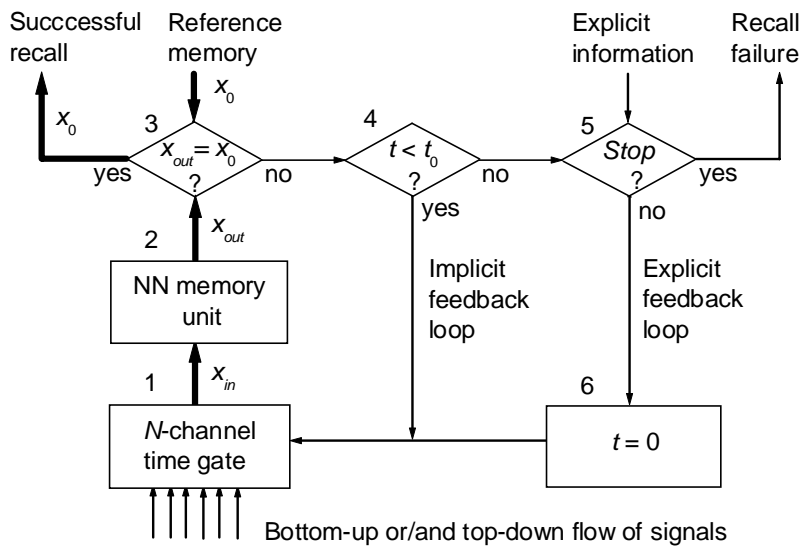

**Рис. 2.** Блок-схема (т.е. архитектура) нейросетевой ансамблевой ячейки памяти и ее ближнее окружение. Пути для распространения групп из $N$ синхронизированных импульсов и связи для передачи управляющих сигналов показаны как толстые и тонкие стрелки, соответственно. На входе *блока 1* импульсы, относящиеся к восходящему или нисходящему потоку спайков, неотличимы. Подробности в тексте.

**4.1. $N$-мерные характеристические векторы.** Поставим в соответствие *$N$-мерному характеристическому вектору $x(d)$* группу из $N$ устойчиво распространяющихся спайков, прецизионно синхронизованных во времени и образующих ярко выраженное событие, которое может быть зарегистрировано чувствительными к совпадениям нейронами даже в других (удаленных) областях мозга. По определению (разд. 1) каждый характеристический вектор $x(d)$ несет долю $q = 1 - d$ истинной информации о произвольной особенности или *характеристике* стимула. Главная задача, которая решается с помощью таких векторов, – это перенос информации между отдельными элементами конкретной ансамблевой ячейки памяти (жирные стрелки на рис. 2) и между разными взаимосвязанными (*ассоциированными*) ячейками ансамблевой памяти, которые могут относиться, например, к определенному стимулу (см. разд. 3 и 5). Поскольку $x(d)$ являются бинарными векторами, то в мозге на любом иерархическом уровне обработки данных они могут переносить в универсальной цифровой форме произвольную информацию, в частности, о внешних или внутренних стимулах, символах или правилах. Это означает, что $x(d)$ может кодировать как отдельную характерную особенность стимула (например, отдельные особенности некоторого образа или события), так и конкретные правила (например, правила для временнόго связывания компонент бистабильных зрительных образов или для решения контекстно-зависимой задачи об отделении фигуры от фона [8, 9]). Характеристические векторы могут интерпретироваться как *сообщения* или *кодовые слова* длины $N$, с помощью которых различные области («устройства») мозга осуществляют обмен информацией между собой.

Линейный хемминговский классификатор вычисляет (разд. 2) свертку $Q(d)$ – проекцию произвольного характеристического вектора $x(d)$ на оптимальный характеристический вектор $x(0)$. Если $Q(d) > 0$ или $Q(d) \leq 0$, то он классифицирует $x(d)$ как искаженный шумом эталонный образ $x_0$ или как образец чистого бинарного шума, соответственно. Оптимальный характеристический вектор $x(0)$ (вектор с максимальным отношением сигнал/шум $q = 1 - d$) и эталонный вектор $x_0$ – это одно и то же, так как в этом случае $x_0 = x(0)$ и $q = 1$ (см. разд. 1).

Оптимальный характеристический вектор $x(0)$ инициирует узнавание, векторы $x(d)$ с $0 < d < 1$ инициируют вспоминание с подсказкой, а векторы $x(1)$, т.е. шум, необходимы для инициирования свободного вспоминания без какой бы то ни было подсказки (см. разд. 7 и 14.2). Это означает, что не только характеристические векторы вызываемые стимулами, но и характеристические векторы не привязанные к конкретным стимулам (случайные образы синхронной нейронной активности, спонтанно возникающие в малых нейронных популяциях [10]) могут играть важную функциональную роль по всему мозгу. Синхронные пакеты спайков, которые переносят (полную) информацию о некоторой особенности конкретного стимула и моделируются (оптимальными) характеристическими векторами, в живых организмах пока не обнаружены (однако см. разд. 13.3 и 13.4). Причина этого, вероятно, в том, что сейчас экспериментально изучают, как правило, только парные корреляции спайковой активности нейронов, а для решения названной задачи нужна техника $N$-клеточных совпадений.

Все модельные спайки, относящиеся к определенному характеристическому вектору $x(d)$, поступают на вход целевой ИНС (блока сравнения или другой ансамблевой ячейки памяти) практически одновременно (рис. 2) и этот факт находится в полном согласии с имеющимися нейробиологическими данными (разд. 13.3). Поскольку векторы $x(d)$ кодированы так (разд. 1), что их алгоритм декодирования обладает свойствами максимального правдоподобия (разд. 2), то естественно предположить, что соответствующие алгоритмы кодирования-декодирования могут реализовывать тот «наилучший нейронный код» [11] или тот «идеальный» кодер-декодер [12], которые исследователи мозга долго и упорно ищут и в теории, и на эксперимнте.

**4.2. Каналы и связи.** Хорошо известно, что в сенсорной системе животных и человека есть параллельные каналы, по которым могут распространяться синхронизованные (или несинхронизованные) сигналы о стимулах. Имеющиеся нейробиологические данные (разд. 13.4) подтверждают также возможность устойчивого распространения групп синхронизованных спайков в очень сложных кортикальных нейронных сетях мозга и, как следствие, поддерживают наше предположение о существовании *путей* (жирные стрелки на рис. 2), вдоль которых могут *устойчиво* распространяться $N$-мерные *характеристические векторы* $x(d)$ (группы синхронизованных спайков). Эти пути для передачи каждой следующей группы спайков на каждом следующем шаге процесса извлечения информации из памяти (например, для передачи каждого следующего образа $x_{in}$, распространяющегося от *блока 1* к *блоку 2* на рис. 2) могут конструироваться всякий раз из другого набора нервных клеток и могут рассматриваться поэтому как *динамически* создаваемые шины *параллельной* (синхронной) передачи данных. По аналогии *связи* (тонкие стрелки на рис. 1) могут рассматриваться как шины *последовательной* передачи данных (асинхронная связь). Поскольку на каждом шаге процесса вспоминания каналы могут динамически создаваться заново, то естественно предположить, что связи могут создаваться тоже динамически.

**4.3. Блоки сравнения (принятия решений).** О свойствах зависящих от времени нейронов и нейронов-детекторов ошибок, которые могут быть существенны при создании *блоков 3,4* и *5*, см. разд. 13.5. Здесь мы только формулируем гипотезу о том, что обычные чувствительные к совпадениям нейроны [13] могут быть более существенно вовлечены в конструирование *блоков 3* и *4*, которые участвуют в импликативной (т.е. неосознаваемой) обработке информации, в то время как нейроны-детекторы ошибок могут более существенно участвовать в создании *блока 5*, который участвует в экспликативной (т.е. осознаваемой) обработке информации (разд. 4.6) и, следовательно, последние могут быть частью *нейронных корреляций сознания* [14, 15]. Основанием для такого предположения служит то, что нейроны-детекторы ошибок изменяют свою активность только тогда, когда делается ошибка при выполнении когнитивных (осознаваемых) действий. В остальном нейроны-детекторы ошибок и обычные чувствительные к совпадениям нейроны обладают, вероятно, близкими свойствами.

Обнаруженные в мозге животных нейроны зависящие от времени (разд. 13.5) идеально подходят для управления модельным процессом импликативного извлечения информации из памяти (*блок 4* и петля 1-2-3-4-1 на рис. 2), поэтому возникает впечатление, что в ходе эволюции они могли быть специально созданы природой для решения (среди прочих) этой весьма специфической задачи.

**4.4. *N*-канальные временные ворота.** С помощью временных ворот (*блок 1* на рис. 2) осуществляется предварительная обработка исходных триарных распределенно кодированных векторов высокой размерности для исключения из текущего рассмотрения дремлющих в данный момент времени нейронов и получения за счет этого достаточно коротких плотно кодированных характеристических векторов $x_{in} = x(d)$ (разд. 4.1). Предполагаем, что группа синхронизованных спайков $x_{in}$, выделенная *блоком 1* из потока восходящих и/или нисходящих входных спайков, может содержать подсказки, обусловленные восприятиями окружающей среды и/или внутренне генерируемыми ожиданиями и/или предчувствиями. *Блок 1* может быть распределенной системой с $N_{sps}$ постоянными входными каналами (это означает, что в живых организмах они задаются отчасти анатомически) и $N_{dns}$ динамически создаваемыми (на каждом шаге извлечения из памяти) выходными каналами, которые являются одновременно $N$ входными каналами ($N = N_{dns}$) нейросетевой ячейки памяти (*блок 2*). Согласно нейрофизиологическим данным значения $N_{sps}$ и $N_{dns}$ можно грубо оценить как 3000-10000 и ~100, соответственно (разд. 13.2, 13.3 и 13.4). Когда к *блоку 1* поступает управляющий сигнал от *блока 4* или *6*, то все $N_{sps}$ входные каналы одновременно открываются на время $\Delta t$, соответствующее ширине временного окна чувствительных к совпадениям нейронов (~1 мс). $N_{dns}$ спайков, которые попадают в это временное окно (все равно обусловлены они стимулами или шумом), и есть та группа из $N = N_{dns}$ синхронизованных сигналов, которая соответствует входному вектору $x_{in}$ для *блока 2*. Таким образом, *блок 1* для *блока 2* действует как *N*-канальные временные ворота с шириною окон $\Delta t$, которые открываются внешним управляющим сигналом.

**4.5. Эталонная память.** Характерной особенностью архитектуры нейросетевой ансамблевой ячейки памяти (рис. 2) является то, что след памяти $x_0$ хранится одновременно в двух разных памятях: в нейросетевой памяти (*блок 2*) и в дополнительной (вспомогательной) *эталонной памяти*, вводимой здесь впервые. Последняя является частью ансамблевой ячейки памяти на этапе ее обучения (или активизации) и частью ее окружения, когда информация извлекается из памяти (как на рис. 2). Эталонная память может интерпретироваться как *ярлык* или *библиографическая карточка* конкретной записи памяти в каталоге библиотеки всех долговременных нейросетевых памятей и выполняет две взаимосвязанные функции: *верификация* текущих результатов вспоминания (*блок 3*) и *подтверждение* (на этапе активизации памяти, см. разд. 5) того факта, что конкретная запрашиваемая запись (след) памяти действительно существует в библиотеке памятей. Это оз начает, что эталонная память может рассматриваться как «память о памяти» или «метапамять» [16]. Другими словами, «память о памяти» есть «знание о памяти». Следовательно, эталонная память является также одним из видов обобщенного представления знаний (в форме $N$-битового бинарного кода) или примером «специфического вида не выражаемого словами знания» [17, с.25].

Подчеркнем разницу между *обычным следом* памяти, хранящимся в обычной ИНС (*блок 2*), и таким же *эталонным следом* памяти из эталонной памяти: обычный след может, а эталонный след не может быть преобразован из активной в пассивную форму и наоборот (разд. 6). Это означает, что обычные следы памяти (векторы или матрицы) участвуют в передаче и обработке информации *непосредственно*, в то время как эталонные следы памяти только *«поддерживают»* эти процессы. В отличие от обычной нейросетевой памяти, которая является разновидностью *компьютерного регистра* (разд. 6) и обычно ассоциируется с реальными нейробиологическими сетями, эталонная память есть некое *гнездо* или *разъем* определенной конфигурации, предназначенный только для *сравнения* различных векторов $x$ с эталоном $x_0$, т.е. этому новому модельному объекту должен соответствовать нейробиологический аналог, природу которого еще предстоит установить.

Различные динамически создаваемые клеточные ансамбли функционально перекрываются и по этой причине один и тот же нейрон может участвовать в процессе обработки разных памятей [18]. При этом в мозге нет какой-либо системы нумерации нейронов, с помощью которой для каждого из них определялся бы его уникальный номер или «адрес», необходимый для конкретизации его роли при хранении, извлечении или передаче информации в непрерывно динамически меняющейся среде активных нейронов. В рамках нашей модели именно эталонный след памяти определяет порядок следования $N$ компонент запомненного следа памяти. Благодаря этому становится возможным извлекать из конкретной ИНС как *количество* разных выходных состояний нейронов (отдельно −1 и отдельно +1), так и *порядок следования* этих состояний. Другими словами, эталонный след памяти обеспечивает для некой популяции нейронов возможность ее синергетического кодирования-декодирования, когда количество хранимой в сети информации (заданная последовательность $N$ нейронных состояний +1 и −1) больше, чем количество информации, которое могут нести $N$ независимых индивидуальных нейронов (полное количество состояний отдельно +1 и отдельно −1).

**4.6. Две вложенные петли обратной связи.** Две вложенные петли обратной связи образуют еще одну отличительную особенность ячейки ансамблевой памяти (рис. 2). Все элементы внутренней или импликативной петли обратной связи (цикл 1-2-3-4-1) выполняются рутинно в автоматическом режиме и по этой причине могут интерпретироваться как относящиеся к *импликативной* (неосознаваемой) памяти. Это означает, согласно модели, что все нейронные операции на уровне синаптических или нейросетевых памятей выполняются в мозге только неосознанно. Внешняя *экспликативная* петля обратной связи (цикл 1-2-3-4-5-6-1) управляется сигналами, поступающими на *блок 5* из непредсказуемо меняющейся внешней среды, и поэтому она активизируется непредсказуемым образом, обеспечивая неограниченное многообразие мод экспликативного извлечения информации из памяти. По этой причине ансамблевая ячейка памяти может рассматриваться как модель *экспликативной* (осознаваемой) памяти и как модель для описания *соотношений* между импликативной и экспликативной памятями. По этой же причине внешняя информация, используемая в *блоке 5*, может интерпретироваться как *экспликативная* или осознаваемая информация (о разнице между импликативной и экспликативной памятями см. [19]), а нейроны-детекторы ошибок, которые согласно модели используются при создании *блока 5*, могут интерпретироваться как относящиеся к *нейронным корреляциям сознания* [14, 15]. Согласно НСАМП только на уровне ансамблевой памяти возникает возможность учитывать экспликативные (осознаваемые) факторы (разд. 14.4) и, следовательно, отдель ная ансамблевая ячейка памяти может рассматриваться как наименьший неделимый элемент или *«атом»* всех возможных осознаваемых памятей. Таким образом, ее можно использовать как строительный материал или «кирпичик» при создании (моделировании) осознаваемых памятей любого уровня или осознаваемых когнитивных функций мозга в целом.

Это утверждение согласуется с представлением о модульной структуре сознания и о множественности составляющих его мелкомасштабных или микро-сознаний, которые были введены в [20, 21] на основе экспериментальных данных, демонстрирующих, что разные зрительные субмодальности (такие как движение или цвет) могут осознаваться с некоторым временным разбросом. Наша модель согласуется также с утверждением о том [22], что сознание может быть тесно связано не с операциями по обработке информации (импликативная петля 1-2-3-4-1), а с их *результатом* (*блок 5*), который затем должен быть распределен по всей системе (петля 1-2-3-4-5-6-1 и другие ассоциированные ансамблевые ячейки памяти). Используя терминологию [22], нашу ансамблевую ячейку памяти можно рассматривать как пример микротеории сознания, так как она имеет дело с аспектами сознания, обусловленными *специфическими процессами* в мозге. Особенности цельной НСАМП как теории сознания мы здесь не обсуждаем, хотя у нее много общих и отличительных черт с такими современными теориями как, например, гипотеза динамического кора [23] или глобального нейронного рабочего пространства [24].

## 5. АКТИВИЗАЦИЯ АНСАМБЛЕВОЙ ПАМЯТИ

Предполагаем, что клеточный ансамбль, который откликается на конкретный стимул и соответствует конкретной ансамблевой памяти о стимуле (мы называем ее «*стимульной*» памятью), выделяется и активизируется (готовится к последующей работе) посредством некоего механизма динамической пространственно-временной синхронизации ([8, 9, 18] и разд. 13.1). Память о стимуле может быть широко распределена по далеко отстоящим и функционально разным областям мозга и может состоять из большого количества взаимосвязанных (*ассоциированных*) «характеристических» ансамблевых ячеек памяти или характеристических памятей, каждая из которых (блок-схема на рис. 2) хранит одну характерную особенность стимула. Памяти о разных характеристиках одного и того же стимула связаны между собой посредством путей и связей (разд. 4.2) как непосредственно, так и при посредничестве сознания и образуют множества и подмножества памятей организованных иерархически, но без соблюдения их строгой иерархической структуры. Если некая особенность стимула представляет текущий интерес (т.е. находится в центре внимания), то будем называть ее *целевой особенностью*. Памяти о стимулах, относящихся к одной и той же категории стимулов, также взаимосвязаны между собой посредством путей и связей, обсуждавшихся в разд. 4.2, как напрямую, так и посредством сознания и тоже образуют множества и подмножества памятей, организованных иерархически, но без соблюдения их строгой иерархической структуры. Если некоторый стимул, представляет текущий интерес (т.е. находится в центре внимания), то будем называть его *целевым стимулом*. Разные клеточные ансамбли, которые соответствуют памятям о разных особенностях одного стимула или памятям о разных стимулах, могут сосуществовать одновременно так, что их взаимное перекрытие не очень велико [18]. Полагают, что разные клеточные ансамбли могут составлять от несколько десятков до нескольких тысяч нейронов [25].

Таким образом, в рамках нашего подхода все памяти, относящиеся к объектам, событиям или правилам, упорядочены во вложенные или частично вложенные множества и подмножества ассоциированных памятей так, что внутри одного множества/подмножества памятей все они равноправны и информация из одной ансамблевой ячейки памяти может использоваться как подсказка для другой такой памяти (это «сеть взаимосвязей одного объекта с другими», [17, с.9]). Если активизируется конкретная *целевая* память (память, представляющая текущий интерес, или, что то же самое, находящаяся в центре внимания), то *вместе* с нею активируются ее эталонная память и все другие функционально связанные с нею ансамблевые памяти из одного множества/подмножества памятей со всеми своими внутренними и внешними путями и связями.

Таким образом, наше рассмотрение предполагает существование нового специфического этапа функционирования памяти, этапа ее *активизации*, когда из огромного хранилища долговременных памятей отбирается (выделяется) и активизируется *одна* целевая память, но *вместе* с ее собственной эталонной памятью, другими ассоциированными памятями, их путями и связями. Причем этот процесс осуществляется с помощью некоего механизма, связанного со вниманием (см., например, [26, 27]). Состояние конкретной памяти, полученное после ее выделения и активизации, отражает ее готовность к дальнейшей работе. Будем называть его *активным состоянием памяти*. Когда активное состояние памяти приготовлено, то целевая память обучается (разд. 10) или тестируется для извлечения хранящегося в ней следа памяти (разд. 4), а все другие активизированные одновременно с нею ассоциативные памяти либо поддерживают этот процесс своими подсказками, либо ожидают момента, когда будут востребованы. Дополнительно полагаем, что целевая память становится составной частью текущей кратковременной (рабочей) памяти. По этой причине целевая память может стать осознаваемой, в то время как все другие ассоциированные с нею и активизированные вместе с нею памяти остаются вне пределов кратковременной памяти и образуют ее неосознаваемое контекстное окружение или «периферию» сознания.

Активное состояние памяти (состояние после ее активизации) можно интерпретировать как возможную количественную модель введенной Е. Тулвингом «моды извлечения из памяти» [28] – специфического когнитивного состояния вспоминающего человека, в котором предоставляемые ему стимулы могут восприниматься как подсказки вспоминания. Кроме того, в рамках нашей модели активному состоянию памяти можно сопоставить аналоги таких специфических когнитивных состояний или процессов как «экфория», взаимодействие между подсказкой вспоминания и следом памяти, настрой на вспоминание, попытка вспоминания, усилие вспоминания и т.д. [29].

Поскольку предполагается, что упомянутые памяти и их ассоциации создаются на стадии их обучения или конструирования (последнее означает, что в живых организмах они задаются отчасти анатомически), то нет необходимости в обычной адресной части сообщений, передаваемых между конкретными ансамблевыми памятями, и, как результат, вообще говоря нет связанных с адресацией ограничений на объем долговрменной (постоянной) памяти, построенной таким способом. Это биологически правдоподобная ситуация, поскольку хорошо известно, что каждое ментальное или поведенческое действие сопровождается активизацией большого количества разных нейронных сетей в различных областях мозга, а емкость долговременной памяти человека может быть огромна (практически она не ограничена).

Здесь и в разд. 12.4 мы обсуждаем композиционные свойства ансамблевых памятей только для того, чтобы показать место отдельной ячейки ансамблевой памяти в цельной картине нейросетевой ансамблевой модели памяти вообще. Обсуждению конкретных способов конструирования композиционных памятей о стимулах, памятей о множествах, или множествах множеств стимулов будет посвящена отдельная работа. То же относится к ансамблевым памятям, хранящим элементарные символьные правила, композиции элементарных символьных правил, их множества или множества их множеств.

# 6. АКТИВНЫЕ И ПАССИВНЫЕ СЛЕДЫ ПАМЯТИ

След, хранящийся в нейросетевой ячейке ансамблевой памяти, может существовать одновременно в активной (вектор $x_0$) и пассивной (матрица $w$) форме (разд. 2).

Синаптическая матрица $w$ – это *пассивный* след памяти. Она хранит информацию в виде специфического распределения сил межнейронных связей (возбудительных, если $w_{ij} > 0$, или замедлительных, если $w_{ij} < 0$). В пассивной форме информация может храниться произвольно долго или даже постоянно, а соответствующая нейронная сеть отчасти моделирует те области или системы мозга, которые связаны с хранением и функционированием конкретной ансамблевой памяти. Ее можно интерпретировать также как *регистр* длины $N$ для обработки бинарных сообщений емкостью $N$ бит.

Эталонный вектор $x_0$ – это *активный* след той же памяти. Векторы $x_0$, $x_{in}$ и $x_{out}$ моделируют синхронизированные группы сигналов, с помощью которых передается информация связанная с памятью, и могут интерпретироваться как ее *количественное представление* в виде $N$-битового *бинарного кода*. В активной форме информация существует только во время распространения группы синхронизованных сигналов (спайков) и, следовательно, должна быть использована в течение этого короткого времени.

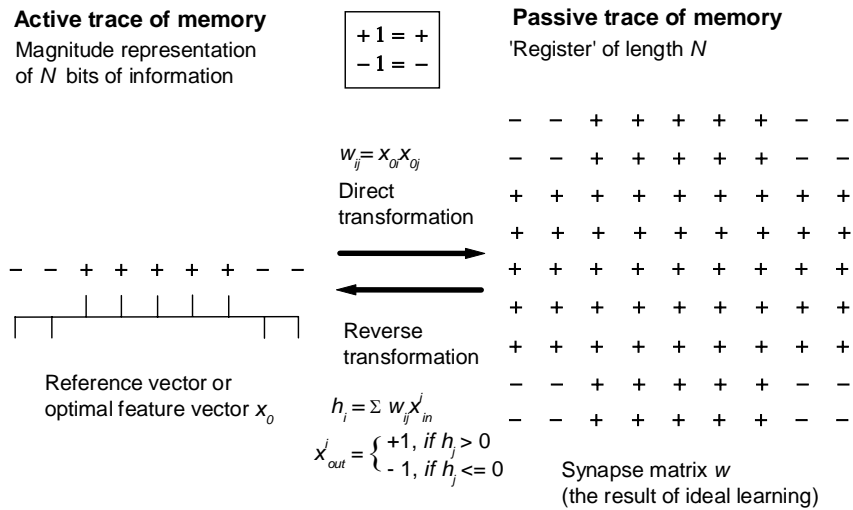

**Рис. 3.** Соотношения между активным и пассивным следами памяти, между их прямым и обратным преобразованиями. Пример идеально обученной неповрежденной ИНС с $N = 9$ (параметр обучения $\eta = 1$). Активный след памяти показан как последовательность положительных и отрицательных единиц и как набор черточек, расположенных выше (+1) или ниже (–1) горизонтальной линии; пассивный след памяти показан как матрица с элементами +1 или –1 (для простоты они обозначены как «+» и «–», соответственно).

Если уравнение (4) из разд. 2 или (10) из разд. 10 постулировать как *прямое* преобразование активного следа памяти в пассивный, то выражения (5) и (6) будут определять их *обратное* преобразование. Более того, нейросетевую ячейку памяти (*блок 2* на рис. 2) удобно рассматривать как *устройство* для преобразования активного следа памяти в пассивный и наоборот (рис.3), а не просто как емкость для хранения информации (следа памяти). Это утверждение находится в полном согласии с популярной точкой зрения о том, что биологически правдоподобные модели должны использовать регистровую архитектуру памяти, а также хранить и манипулировать математически с информацией, представленной количественно [30].

Тот факт, что активные и пассивные следы памяти являются квазибинарными, означает, что вычисления (обработка информации) в нервных тканях животных и человека выполняются, согласно НСАМП, в квазибинарной арифметике.

## 7. МАКСИМАЛЬНО-ПРАВДОПОДОБНЫЕ ХАРАКТЕРИСТИКИ ПАМЯТИ

Считаем, что ИНС, обученная как это описано в разд. 2, успешно распознает $x_0$ в $x_{in}$, кодированном как это описано в разд. 1, если отклик сети $x_{out}$ на $x_{in}$ есть $x_0$. С другой стороны, та же ИНС есть нейросетевая ячейка памяти и выражения (5) и (6), или неравенство (7), являются частью алгоритма извлечения информации из памяти, который предполагает тестирование сети серией разных векторов $x_{in} = x(d)$, которые генерируются случайно, но с сохранением постоянного значения подсказки $q = 1 - d$. Если мы определим вероятность узнавания $x_0$ в $x_{in} = x(d)$ как $P(d)$, то $P(d)$ есть одновременно вероятность вспоминания (извлечения из памяти) следа $x_0$ сетью, тестируемой серией векторов $x_{in}$ с постоянной подсказкой $q$. Входной вектор $x_{in} = x_0$ распознается идеально обученной неповрежденной сетью или сетью с некатастрофическими повреждениями как эталон $x_0$ с вероятностью $P(0) = 1$; шум $x_{in} = x_r$ интерпретируется как $x_0$ случайно с вероятностью $P(1) = \alpha < 1$. В результате имеем $\alpha \leq P(d) \leq 1$.

Ниже будем рассматривать введенный алгоритм распознавания или вспоминания (извлечения из памяти) как *критерий* для различения простых альтернативных статистических гипотез.

Пусть *нулевая гипотеза* $H_0$ состоит в том, что статистический образец $x_{in} = x(d)$ есть образец шума; в этом случае $x_{in} = x_r$ и $d = 1$. По определению [31] вероятность $\alpha = P(1)$ отвергнуть $H_0$ есть *уровень статистической значимости* критерия а также *вероятность ошибки первого рода* и *вероятность ложного обнаружения*. Шум $x_r$ не содержит никакой информации об $x_0$ и, следовательно, $\alpha$ есть также вероятность вспоминания $x_0$ без какой либо подсказки или *вероятность свободного вспоминания*. Следовательно, $\alpha$ есть одновременно *уровень значимости критерия, вероятность ошибки первого рода, условная вероятность ложного обнаружения (вероятность ложной тревоги)*, и *вероятность свободного вспоминания*.

*Альтернативная гипотеза* $H_1$ состоит в том, что статистический образец $x_{in}$ есть $x_0$ со степенью его искажения $d$; в этом случае $x_{in} = x(d)$ и $0 \leq d < 1$. Если вероятность отвергнуть $H_1$, когда она верна, есть $\beta$, то $\beta$ есть *вероятность ошибки второго рода*. Вероятность принятия $H_1$ при том же условии есть *мощность критерия* $M = 1 - \beta$. Кроме того, поскольку образец $x(d)$ вызывает отклик сети $x_0$ с вероятностью $P(d)$, то $M = P(d)$. Если статистический образец $x(d)$ содержит часть $q$ информации об $x_0$, то она «напоминает» обученной ИНС об $x_0$ и таким способом «помогает» его вспомнить. Следовательно, $P(d)$ есть также вероятность вспоминания $x_0$ с подсказкой $q = 1 - d$. Поэтому $P(d)$ – это одновременно *мощность критерия, условная вероятность истинного обнаружения*, и *вероятность вспоминания с подсказкой* $q = 1 - d$. Тогда $P(0)$, частный случай $P(d)$ при $d = 0$, – это *вероятность узнавания*.

Подчеркнем, что $P(d)$ – это условные вероятности. Условия, при которых они найдены, – это справедливость либо гипотезы $H_0$, либо гипотезы $H_1$. Однако с помощью известной формулы Байеса можно найти [3] *безусловную вероятность ложной классификации* (*ложного вспоминания или обнаружения*)

$$P_{MC} = 1/(1 + \kappa P(d)/P(1)) \tag{8}$$

и *безусловную вероятность правильной классификации* (*правильного вспоминания или обнаружения*)

$$P_{CC} = 1/(1 + \kappa^{-1} P(1)/P(d)). \tag{9}$$

Отношение $\kappa = P(H_1)/P(H_0)$, где $P(H_0)$ и $P(H_1)$ – априорные вероятности справедливости гипотез $H_0$ и $H_1$, учитывает ту дополнительную априорную информацию, которая существенна для правильного декодирования, но отсутствует в анализируемых данных. Если *предположить*, что $0 < \kappa < 1$ и $\kappa = 1$, то $0 < P_{MC} \leq \frac{1}{2}$ и $\frac{1}{2} \leq P_{CC} < 1$, $P_{MC} + P_{CC} = 1$ (некоторые подробности см. [3]). Для заданного алгоритма декодирования (вспоминания, узнавания, обнаружения или классификации) зависимость условной вероятности правильного обнаружения $P(d)$ от значения условной вероятности ложного обнаружения $\alpha = P(1)$ называют кривой операционных характеристик приемника или кривой ОХП (см., например, [32] и разд. 11.4). Из приведенных выше неравенств, ограничивающих значения $P_{MC}$ и $P_{CC}$, следует, что кривые ОХП могут быть либо линейными, либо выпуклыми.

Наше рассмотрение показывает, что в рамках НСАМП широко используемые в нейропсихологии [33] *основные характеристики памяти* (вероятности свободного вспоминания, вспоминания с подсказкой и узнавания) получают строгое количественное определение и теперь могут быть вычислены точно [1]. Для введенного нами способа кодирования функции $P(d)$ описывают показатели качества и алгоритма декодирования-обнаружения, и алгоритма вспоминания-узнавания, поэтому в рамках нашего подхода и те, и другие являются максимально-правдоподобными; в частности, максимально правдоподобными являются основные характеристики памяти. Кроме того, для любой памяти, все равно относится она к живому организму или машине, вероятности $P(d) = P(1 - q)$ могут быть определены эмпирически путем анализа достаточно большого набора соответствующих статистических образцов [3].

Правила и формулы для вычисления $P(d) = P(1 - q)$, а также численные примеры см. в разделе 11.

## 8. ЗАКРЕПЛЕНИЕ И ОСЛАБЛЕНИЕ АНСАМБЛЕВОЙ ПАМЯТИ

Закрепление ансамблевой памяти предполагает закрепление следа, хранящегося в нейросетевой ячейке памяти (*блок 2* на рис. 2), закрепление эталонного следа памяти и всех относящихся к ним путей и связей. По этой причине моделирование ослаблений ансамблевой памяти предполагает возможность повреждения всех названных ее компонентов. Ниже мы обсудим дефекты памяти, обусловленные только повреждениями ИНС из состава ансамблевой ячейки памяти, полагая, что эталонная память, а также пути и связи, показанные на рис. 2, были найдены как результат идеального обучения и сохранены полностью.

Идеально обученная ИНС может быть частично повреждена в результате потери части ее входных нейронов, нейронных связей между входными и выходными нейронами и путем частичной стохастизации сил межнейронных синаптических связей. Все эти факторы могут действовать как вместе, так и каждый в отдельности. Если ослабление памяти обусловлено естественными причинами, то это обычное *забывание*, в противном случае это результат *болезни* мозга или его *травмы*. Рис. 4 в разд. 11 демонстрирует как ослабление памяти (уменьшение значений $P(d)$ при фиксированном $d$) зависит от распределений нескольких погибших входных нейронов вместе со всеми их связями (кривые 3 и 4) или от распределений нескольких разорванных связей между входными и выходными нейронами сети (кривые 2, 5 и 6). При этом предполагаем, что для каждой $ij$-связи, *разорванной* между *i*-м входным и *j*-м выходным нейронами, соответствующий элемент синаптической матрицы сети есть $w_{ij} = 0$. Для каждого *i*-го *погибшего* входного нейрона полагаем, что разорваны его связи со *всеми* выходными нейронами сети и поэтому $w_{ij} = 0$ для заданного *i* и всех $j = 1,..,N$. Если есть погибший нейрон в *выходном* слое сети, то соответствующая нейросетевая ячейка памяти становится полностью неработоспособной, так как размерность $N_{out}$ ее выходного вектора $x_{out}$ становится меньше размерности $N$ соответствующего эталонного вектора $x_0$ и их побитовое сравнение в *блоке 3* (рис. 2) становится невозможным. Следовательно, наличие погибших выходных нейронов в выходном слое обученной ИНС приводят к полной потере работоспособности соответствующей ей ячейки памяти.

Согласно уравнению (4) элементы синаптической матрицы $w_{ij}$ для идеально обученной неповрежденной ИНС определяются вектором $x_0$, который не содержит нулевых компонент. По этой причине для неповрежденной сети все значения $w_{ij}$ ненулевые и, благодаря этому, в рамках НСАМП появляется дополнительный свободный параметр – возможность положить для любой *ij*-связи $w_{ij} = 0$. Это обстоятельство (связанное с тем, что векторы *x* являются не бинарными, а *квазибинарными* и имеют еще третью, нулевую, проекцию) и было использовано выше для определения локальных повреждений модельной нейросетевой памяти (сравни с разд. 1).

## 9. СВОБОДНЫЕ ПАРАМЕТРЫ МОДЕЛИ

Ниже перечислены свободные параметры нейросетевой ансамблевой модели памяти, необходимые для подгонки ее количественных предсказаний к реальным экспериментальным данным:

- *Длина N* нейросетевой ячейки памяти (и соответствующей ей эталонной памяти), равная ее информационной емкости в битах и размерности векторов $x(d)$, которые обслуживают эту память. Поскольку любые сообщения всегда содержат конечное количество информации, то $N$ – это конечное и достаточно малое число, а предел $N \to \infty$ не представляет существенного интереса; биологически правдоподобное значение $N$ около 100 (см. разд. 13.2, 13.3 и 13.4).
- *Частота f,* или период $1/f$, с которой нейросетевая ячейка памяти тестируется векторами $x_{in}$ при извлечении следа памяти $x_0$ (см. разд. 4 и рис. 2). На основании существующих биологически правдоподобных аргументов полагаем, что значения *f* относятся к так называемой гамма-полосе электрической активности мозга, которая составляет ~ 40 Гц (см. разд. 13.1 и 14.3). Это означает, что модельная частота деполяризации активных нейронов является относительно небольшой, в соответствии с реальной нейробиологической ситуацией и в отличие от распределенно кодированных сетей, активные нейроны которых разряжаются с максимальной частотой [6].
- *Максимальная продолжительность* процесса импликативного извлечения информации из памяти $t_0$ или максимальное число $ft_0$ векторов $x_{in}$, которые могут быть использованы при импликативном извлечении из памяти (петля 1-2-3-4-1 на рис. 2). Полагаем, что $t_0$ это характерное время зависящих от времени нейронов и, следовательно, оно может быть из диапазона 10 мс – 10 с (см. разд. 13.5).
- *Число и распределение разорванных связей, число и распределение погибших входных нейронов*, используемых при моделировании ослаблений нейросетевой памяти, вызываемых естественным забыванием или локальными повреждениями мозга (см. разд. 8).

- *Степень стохастизации* элементов синаптической матрицы $w_{ij}$ при моделировании нелокальных повреждений мозга.

- *Значение обучающего параметра η* в уравнении (10) разд. 10 для моделирования процесса неидеального обучения (в (4) конкретное значение η > 0 не имеет существенного значения, так как это уравнение описывает случай идеального обучения).
- *Внешний сигнал «остановиться-продолжать»* на входе *блока 5* (рис. 2), который указывает, следует или нет заново начинать счет времени в *блоке 6* и повторять цикл 1-2-3-4-1 импликативного извлечения из памяти, если время $t_0$ вышло, а положительный результат вспоминания все еще не достигнут. Поскольку у конкретной ячейки ансамблевой памяти нет внутренних механизмов принятия требуемого решения, то полагаем, что оно обусловлено состоянием окружающей среды и формируется осознанно.

## 10. ОДНОШАГОВОЕ ОБУЧЕНИЕ

Уравнение (4) определяет идеальные силы (или степени *желательности*) межнейронных связей $w_{ij}$, так как они непосредственно вычисляются по точному правилу на основе точной информации об $x_0$. Такое правило можно интерпретировать как некую процедуру одношагового обучения *с «супервизором»* $x_0$. Но во многих случаях нужна процедура обучения *без супервизора*.

Рассмотрим традиционное хеббовское обучение согласно *дельта правилу*, которое запишем в виде

$$w_{ij}^{(n+1)} = w_{ij}^{(n)} + \eta \, v_j^{(n)} h_i^{(n)}, \tag{10}$$

где $n$ – номер итерации, а другие обозначения см. в разд. 2. Если $\eta$ мало ($\eta < 1$), то скорость обучения также мала и асимпотические значения $w_{ij}$, определяемые уравнением (4), не достигаются. Если $\eta$ велико ($\eta > 100$), то итерационный процесс приводит к быстрому одношаговому обучению без «катастрофического забывания», так как уже первая итерация дает результат близкий к асимптоте и следующие итерации не ведут к существенному улучшению результата. Мы приведем пример, когда $N = 40$, для $w_{ij}, v_j, h_i, x^i_{in}$ и $x^i_{out}$ разрешены непрерывные значения, а все исходные значения $w_{ij}$ выбираются из диапазона $[-1,1]$ случайно и равновероятно. Если в качестве обучающего образа выбрать $x_{in} = x_0$, то выход сети $x_{out}$ есть приближенная оценка $x_0$. Если $\eta = 400$, то уже первая итерация по дельта правилу дает оценку качества этого приближения $\sum_i |x^i_{out} - x^i_0| < 10^{-30}$. Это более чем достаточно для практических применений.

Одновременно с обучением нейросетевой ячейки памяти создаются соответствующая ей эталонная память и все относящиеся к ним пути и связи. Но конкретные правила для их конструирования мы здесь не обсуждаем.

## 11. ЧИСЛЕННЫЙ ПРИМЕР

**11.1. Определение вероятностей $P(d)$ путем многократных вычислений.** Чтобы подчеркнуть дискретный характер модельных результатов и облегчить их сопоставление с результатами наших предшествующих работ [1, 4, 67, 68], рассмотрим ИНС с малым числом $N = 9$. Условные вероятности вспоминания-узнавания $P(d)$, введенные в разд. 7, будем вычислять как $P(d) = n(d)/n_0(d)$, где $n_0(d)$ – число разных случайно генерируемых входных векторов $x_{in} = x(d)$ с постоянным значением $d$, $n(d)$ – число таких входных векторов, которые ведут к всплыванию на выходе ИНС отклика $x_{out} = x_0$ (эталонный вектор $x_0$ имеет заданную последовательность его компонент ±1). При малых $N$ $P(d)$ можно легко вычислить точно (см. примеры на рис. 4) так как $n_0(d)$ мало и все возможные входы могут быть учтены: $n_0(d) = 2^m C^N_m$, где $d = m/N$ и $m \leq N$ есть число меток $u_i = 1$ в уравнении (2). При боль ших $N$ $P(d)$ можно вычислить путем таких многократных вычислений приближенно, но с любой наперед заданной точностью [1].

**11.2. Аналитические формулы для $P(d)$.** Для неповрежденной идеально обученной ИНС неравество (7) позволяет получить простые аналитические формулы (табл. 1) для вычисления точных максимально-правдоподобных значений $P(d)$ при $d = m/N$ или $P(q)$ при $q = 1 - m/N$, которые для $N = 9$ показаны в виде незаполненных кружков вдоль кривой 1 на рис. 4. В частности, из таблицы видно, что если $d \leq (N - 1)/(2N)$ и $N$ нечетно или $d \leq (N/2 - 1)/N$ и $N$ четно, то $P(d) = 1$; при $d = 1$ $P(1) = \frac{1}{2}$, если $N$ нечетно, и $P(1) = \frac{1}{2} - C^N_{N/2}/2^{N+1}$, если $N$ четно ($C^N_{N/2}/2^{N+1}$ – это вероятность того, что свертка $Q(d)$ при $d = 1$ обращается в нуль). В правой стороне неравенств, определяющих значения $d$, при которых $P(d) = 1$, знак минус выбран потому, что в выражении (6) случай $h_j = 0$ принимают во внимание вместе со случаями $h_j < 0$ (иначе следует выбирать знак плюс). По той же причине выбираем знак минус перед вторым членом в формуле для $P(1)$ при $N$ четных (иначе следует выбирать знак плюс и тогда $P(1)$ будет больше $\frac{1}{2}$). Разности значений $P(1)$ между ближайшими четным и нечетным $N$ можно вычислить, используя соотношение $C^N_{N/2}/2^{N+1}$ ~ $0.4/\sqrt{N}$, найденного с помощью формулы Стирлинга:.

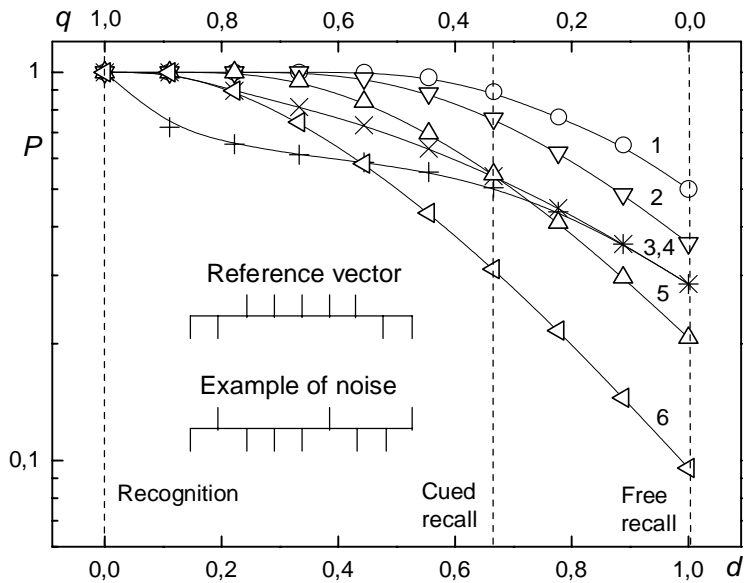

**Рис. 4.** Вероятность $P$ как функция $d$ или $q$ для идеально обученной неповрежденной (кривая 1) и поврежденных (кривые 2-6) ИНС с $N = 9$. Результаты точных вычислений (разные значки) соединены гладкими интерполяционными кривыми. Для кривых 3 и 4 число «погибших» входных нейронов сети $N_k = 4$; для кривых 2, 5 и 6 число разорванных связей между входными и выходными нейронами $N_d = 10$. Штриховые линии отмечают вероятности свободного вспоминания ($q = 0$), вспоминания с подсказкой ($q = 1/3$) и узнавания ($q = 1$). На вставке компоненты $\pm 1$ векторов эталона и примера шума показаны черточками, как на рис. 1 и 3.

Для ИНС с несколькими случайно разорванными межнейронными связями и/или несколькими случайно погибшими входными нейронами максимально-правдоподобные значения $P$ (значки вдоль кривых 2-6 на рис. 4) могут быть найдены путем многократных вычислений (см. выше) или по формулам (здесь они не приводятся), полученным отдельно для каждого конкретного случая локального повреждения сети. Если синаптические матричные элементы сети отчасти стохастизированы, то такие сети не обладают максимально-правдоподобными свойствами вспоминания-узнавания и в этом случае точные аналитические формулы для вычисления $P(d)$ получить нельзя.

Формулы в табл. 1 были получены при условии, что порог срабатывания модельных нейронов $l = 0$ (разд. 2). Если это не так, то конкретный вид формул для $P(d)$ зависит от выбора значения $l$. Например, для нечетных $N$ и $d = 1$ при $l > 0$ и $l < 0$ вместо $P(1) = \frac{1}{2}$ получим, соответственно, $P(1) < \frac{1}{2}$ или $P(1) > \frac{1}{2}$. Следовательно, $l$ может нести ту априорную информацию, которая отсутствует в анализируемых данных, но существенно влияет на результат их анализа.

**Таблица 1.**
Аналитические формулы для точного максимально правдоподобного вычисления значений $P(m,N)$ = $P(d)$ = $P(1-q)$ вероятностей вспоминания или узнавания следа памяти $x_0$ в зависимости от длины $N$ идеально обученной неповрежденной ИНС и числа $m$ искаженнных шумом компонент характеристических векторов $x(d)$; $d = m/N$, $q = 1 - m/N$.

| $N$ is odd | | $N$ is even | |
|---|---|---|---|
| $m$ | $P(m,N)$ | $m$ | $P(m,N)$ |
| $0 \leq m \leq (N-1)/2$ | 1 | $0 \leq m \leq N/2 - 1$ | 1 |
| $(N-1)/2 < m \leq N$ | $\sum_{i=0}^{(N-1)/2} C_i^m / 2^m$ | $N/2 - 1 < m \leq N$ | $\sum_{i=0}^{N/2-1} C_i^m / 2^m$ |
| $N$ | ½ | $N$ | $1/2 - C_{N/2}^N / 2^{N+1}$ |

**11.3.** $P(d)$ **как обобщающая способность сети.** Для используемых двухслойных автоассоционных ИНС вероятности $P(d)$ можно также интерпретировать как их максимально-правдоподобную *обобщающую способность*. Например, если $d = 1$, то неповрежденная ИНС (см. кривую 1 на рис. 2) дает отклик $x_{out} = x_0$ для половины всех возможных входов $x_{in} = x_r$, т.е. в этом случае обобщающая способность сети соответствует случайному угадыванию ответа, $P(1) = ½$. Но для значений $d = m/N$, взятых из соответствующей строки табл. 1, декодирование является уже идеальным, так как в этом случае обобщающая способность сети есть $P(d) = 1$ и все анализируемые входы дают правильный результат вспоминания.

**11.4. Соотношение между** $P(d)$ **и кривыми ОХП.** При анализе результатов измерений, относящихся к исследованию нейронных кодов используемых живыми организмами для кодирования-декодирования сенсорной информации, широко используют кривые операционных характеристик приемника (ОХП) (см. [32, 34] и разд. 7). Поэтому на рис. 5 некоторая часть количественных данных рис. 4 представлена в форме кривых ОХП, привычных для нейрофизиологов. Обычные кривые ОХП и кривые на рис. 5 выглядят похоже, но они совершенно разные. Действительно, в обоих случаях для соответствующих кривых статистическая обеспеченность анализируемых данных одинакова (одинаково значение отношения сигнал/шум $q = 1 - d$), но кривые ОХП представляют $P(d)$ как функцию $P(1)$ для *одного и того же* алгоритма декодирования, а кривые на рис. 5 соединяют значения $P(d)$ для пар значений $P(d)$ и $P(1)$, взятых из рис. 4 и относящихся к *разным*, выбранным практически случайно, алгоритмам декодирования, вспоминания или узнавания.

Если предположить, что порог $l$ срабатывания модельных нейронов меняется (см. [4] и раздел 2), то каждая отдельная функция $P(d)$ на рис. 4 (например, кривая 1) будет расщепляться на *семейство* похожих функций, но с разными значениями $P(1)$. Полученное семейство функций $P(d)$, теперь уже относящихся к одному алгоритму декодирования, можно преобразовать в семейство кривых ОХП, каждая из которых будет соответствовать конкретному значению отношения сигнал/шум $q$ или, что то же самое, определенному качеству (статистической обеспеченности) анализируемых данных. Следовательно, кривые ОХП и функции $P(d)$ являются разными формами представления одних и тех же качественных характеристик для одного и того же алгоритма декодироования, обнаружения, вспоминания, узнавания или классификации. Однако $P(d)$, соответствующие поврежденным сетям с погибшими входными нейронами (например, кривые 3 и 4 на рис. 4), должны рассматриваться как специальный случай, так как соответствующие им кривые ОХП имеют нетрадиционную форму, которая до сих пор не была известна ни в теории, ни в эксперименте [32, 34].

**11.5.** $P(d)$ **и объяснение зеркального эффекта памяти.** Одна из основных нерешенных проблем современных моделей памяти – это так называемый зеркальный эффект [35], состоящий в том, что если стимул **A** (объект или событие) запомнен лучше, чем стимул **B**, то частота «попаданий» (вероятность правильных ответов «да» при предъявлении испытуемому истинного стимула) для **A** выше чем для **B**, но, в то же время, вероятность правильных отклонений (вероятность правильных ответов «нет» при предъявлении испытуемому ложного стимула) для **B** выше чем для **A**. В терминах вероятностей вспоминания то же по определению означает, что $P_A(d) > P_B(d)$ при $d < 1$ и $1 - P_A(d) < 1 - P_B(d)$ при $d = 1$, где $P_A(d)$ и $P_B(d)$ есть вероятности вспоминания для стимулов **A** и **B**, соответственно (мы можем предположить, например, что на рис. 4 $P_A(d)$ соответствует кривая 1, а $P_B(d)$ соответствует кривая 6). Отсюда следует (см. рис. 4), что известный из эксперимента зеркальный эффект памяти можно рассматривать как прямое следствие НСАМП.

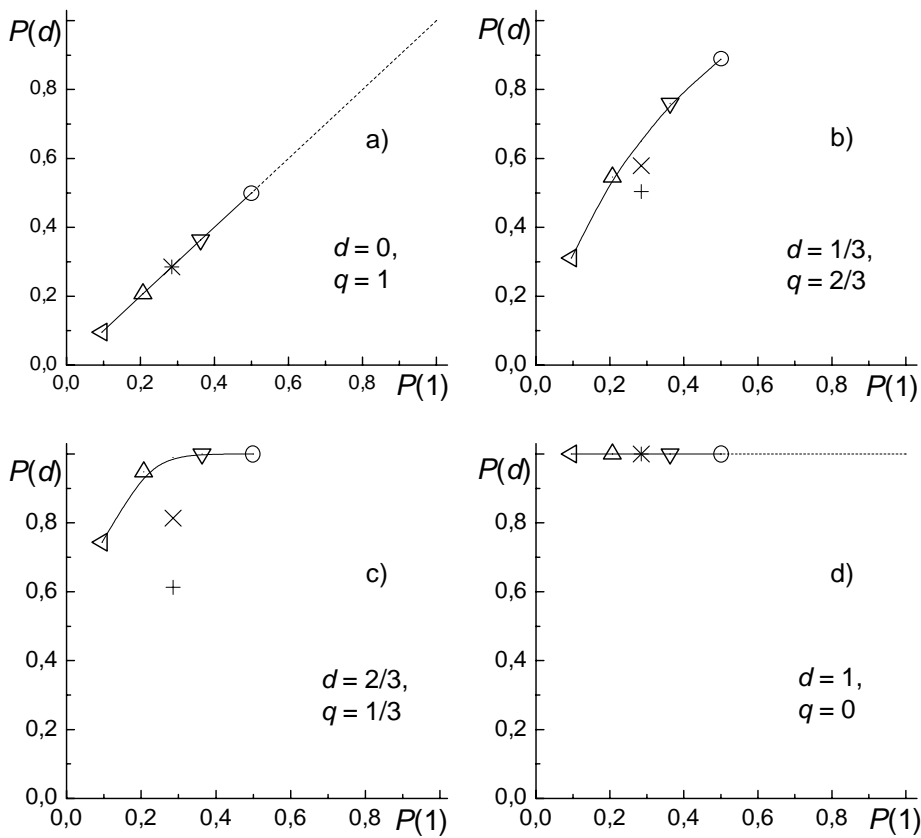

**Рис. 5.** То же, что на рис. 4, но в координатах, принятых для операционных характеристик приемника (ОХП): условная вероятность правильного вспоминания $P(d)$ в зависимости от условной вероятности ложного вспоминания (ложной тревоги) $P(1)$. Каждый значок представляет те же значения $P(d)$, что и на рис. 4: кружки – неповрежденная ИНС; разные треугольники – ИНС с разными распределениями $N_d = 10$ разорванных связей между нейронами входного и выходного слоя; разные крестики – ИНС с разными распределениями $N_k = 4$ погибших входных нейронов. Поскольку в нашем рассмотрении порог срабатывания модельных нейронов $l = 0$, то значения $P(1) > 0.5$ невозможны и область ½ < $P(d) \leq 1$ на рисунке пустая. Штриховые линии на фрагментах a) и d) представляют ожидаемые значения $P(d)$ для нейросетевых алгоритмов вспоминания с $l < 0$. Гладкие интерполяционные кривые соединяют значения $P(d)$ для неповрежденной ИНС и ИНС с разорванными связями, значения $P(d)$ для ИНС с погибшими входными нейронами не ложатся на эти кривые (кроме предельных случаев $d = 0$ и $d = 1$). Каждый фрагмент рисунка показывает значения $P(d)$, которые соответствуют определенному значению $d$ или $q = 1 - d$: a) $d = 0$, b) $d = 1/3$, c) $d = 2/3$ и d) $d = 1$ (значения, показанные для $d = 0, 2/3,$ и 1, на рис. 4 находятся на пересечениях кривых $P(d)$ и соответствующих штриховых линий).

# 12. СРАВНЕНИЕ С ДРУГИМИ МОДЕЛЯМИ ПАМЯТИ

**12.1. Хопфилдовские нейросетевые модели памяти.** Для ИНС типа Хопфилда максимальное число запоминаемых ею образов есть $\alpha N$ ($\alpha \approx 0.14$), а среднее число активных нейронов $F = \frac{1}{2}$ (т.е. при запоминании каждого образа приблизительно половина нейронов сети являются активными). При полном заполнении такой нейросетевой памяти ее информационная емкость есть $I_c = \alpha S(F)$, где $S(F) = -(F\ln(F) + (1-F)\ln(1-F))/\ln 2$ обозначает информационную емкость в битах на одну связь запоминаемого образа с активностью $F$. С помощью хопфилдовских ИНС образы, содержащие поля неизвестных значений, доукомплектовают. Для этого исходный некомплектный образ (вектор) $x_{in}$ подают на вход ИНС, которая преобразовывает его в выходной образ $x_{out}$. Результат этого первого «прохода» через сеть становится входом для следующего такого прохода, для чего полагают $x_{in} = x_{out}$, и т.д. После каждого следующего прохода через сеть каждый следующий образ $x_{out}$ содержит меньше шума, чем предыдущий, поэтому не содержащий шума итоговый образ $x$ получают из $x_{in}$ после ряда итераций (проходов), количество которых может быть 30-50. При этом поля исходного образа $x_{in}$ с известными значениями обычно «связывают», т.е. их оставляют неизменными от одного прохода через сеть к другому даже тогда, когда они не согласуются с очередным выходным образом $x_{out}$. Такую сеть называют *памятью адресуемой по содержанию*. Она имеет тенденцию работать недостаточно хорошо, когда доля шума в исходном образе достигает 40% [2, 7].

Наша модель использует элементы синаптической матрицы, вычисляемые согласно уравнению (4), но в отличие от хопфилдовских сетей [2] она предполагает, что каждая нейросетевая ячейка памяти обучена хранить *один и только один* образ, что *все* модельные нейроны сети активны ($F = 1$), а ее информационная емкость на одну связь в битах составляет $1/N$ для неповрежденных ИНС и более $1/N$ для сетей с локальными повреждениями. Предлагаемый механизм извлечения из памяти также является многошаговым (см. разд. 4), но для конкретной ИНС ее входы $x_{in}$ не являются объектом доукомплектования: они генерируются случайно при постоянном значении $d = 1 - q$ и только *инициируют* вспоминание. В нашем случае надо просто *ждать*, когда на очередном (случайном) шаге вспоминания входной вектор $x_{in}$ приведет к всплыванию на выходе сети вектора $x_{out} = x_0$ сразу и целиком с вероятностью $P(d)$, определяемой как описано в разд. 7 и 11. Следовательно, в рамках НСАМП проблема слишком большого количества шума во входных данных сети вообще не возникает, так как даже чистый шум может быть использован для инициализации успешного вспоминания (случай $d = 1$ на рис. 4). Поскольку все векторы входной последовательности $x_{in}$ содержат постоянную часть $q$ неискаженных компонент, то этот факт можно интерпретировать как аналог обычному в нейросетях связыванию. Предложенное (разд. 11.3) определение обобщающей способности нейросетей также существенно отличается от того, которое используют в рамках обычного статистического подхода, так как в нашем случае длина сети $N$ существенно ограничена количеством хранимой (переносимой) информации, а размер используемого обучающего множества есть $k = 1$, т.е. оно всегда состоит только из одного элемента – это $x_0$. Более того, в нашем случае обобщающая способность зависит от подсказки $q = 1 - d$, а не от относительного размера обучающего множесва $k/N$. В результате, если $d = 0$, то даже для поврежденных сетей обобщающая способность $P(d)$ может быть идеальной, т.е. $P(d) = 1$ (обычно обобщающая способность только стремится к единице, если $k/N \to \infty$). Наконец, если в процессе вспоминания вход $x_{in}$ вызывает отклик сети $x_{out} \neq x_0$, то в рамках НСАМП этот результат интерпретируется как потенциально информативный, но не соответствующий ситуации случай, а не как ошибка.

**12.2. Распределенно кодированные нейросетевые памяти.** Полностью распределенное, «разбросанное» или «разреженное» кодирование обосновывают тем, что среди огромного числа нейронов нервной системы только очень малая их часть активна в каждый данный момент времени. Хотя с другой стороны, такие коды не в состоянии описывать существующие одновременно множественные динамические факты или правила и не согласуются с фактом существования отдельных локальных областей мозга, которые ответственны за его разные специфические функции.

Для распределенно кодированных памятей их исходные данные (образы, кодовые слова или векторы) содержат относительно небольшую долю информативных компонент. Такие векторы представляют собой длинные, чаще всего бинарные (0,1), последовательности длины $N$, число единиц в которых невелико. Например, разбросанно кодируемая память Канервы [36] запоминает все хранимые образы в обычной нейросетевой памяти адресуемой по содержанию, а каждое поле запоминаемого образа кодируется путем побитового комбинирования его «полевых» и «значущих» идентификаторов посредством операции булевского исключающего ИЛИ (XOR); последовательности, полученные в результате такого «связывания», тоже имеют длину $N$. Далее все поля, закодированные посредством связывания, комбинируют в единый образ длины $N$ с использованием порогового правила побитового старшинства (это так называемое «разбиение»: например, если $K$ нечетно, то сумма $K$ образов сравнивается с порогом $K/2$). Декодирование (извлечение составляющих образов из составного) также выполняют с помощью операции XOR, так как она является обратной функцией по отношению к себе самой (это так называемый «зондаж»). Результат зондажа может содержать 30-40% шума и поэтому его «очищают» в памяти адресуемой по содержанию или «очистительной» памяти, которая доукомплектовывает получаемый образ. В итоге после многократных проходов через очистительную память может быть извлечено *несколько* образов, ближайших к искомому.

Наша нейросетевая ансамблевая память является и распределенно кодированной, и плотно кодированной *одновременно* (см. рис. 1). Действительно, ее исходные векторы являются распределенно (разбросанно или разреженно) кодированными, т.е., как и в случае модели Канервы, они содержат много случайно распределенных неинформативных (нулевых) компонент вместе с относительно небольшим количеством информативных (±1) компонент. Но после предобработки исходных векторов (*блок 1* на рис. 2) ячейка нейросетевой памяти (*блок 2*) имеет дело, в отличие от модели Канервы, с относительно короткими плотно кодированными векторами, так как исходные триарные векторы к этому моменту уже преобразованы в квазибинарные векторы, содержащие только ненулевые компоненты (см. разд. 1 и 4.4). Благодаря такой дуальной природе НСАМП, она в большей мере биологически правдоподобна и вычислительно простая. Связывание и разбиение, зондаж и очистка модели Канервы являются аналогами, соответственно, кодирования (запоминания сетью следа $x_0$) и декодирования (извлечения из обученной сети вектора $x_{out} = x_0$) нашей модели. В дополнение к тому преимуществу, что наш алгоритм кодирования-декодирования является максимально-правдоподобным, он дает также *только один* свободный от фона результирующий образ $x_0$ вместо нескольких образов, ближайших к искомому, которые в модели Канервы требуют еще дополнительной процедуры верификации для подтверждения того, что из очистительной памяти извлечен именно нужный (а не другой, но близкий к требуемому) образ. Достоинство НСАМП также в том, что ее активные нейроны разряжаются с относительно низкой средней частотой, о чем уже упоминалось в разд. 9.

**12.3. Нейросетевые памяти типа свертки.** Конволюционные (типа свертки) алгоритмы популярны и при решении задач распознавания, и при моделировании памяти. Конволюционные алгоритмы распознавания применяют, чтобы улучшить извлечение элементарных локальных особенностей из цельных одно- или двумерных изображений (например, из рукописного текста или произносимой фразы). Они используют *локальные рецептивные поля* (множества близкорасположенных элементов изображения) и *детекторы локальных особенностей* (специфически организованные множества близкорасположен ных нейронов для анализа рецептивных полей). В разных местах изображения выходы детекторов особенностей с разными *корами извлечения особенностей* идентифицируют разные локальные особенности (например, ориентированные края, оконечности линий или углы), которые образуют *карту особенностей* изображения, определяющую конкретный распознаваемый объект (например, букву или цифру). Упомянутые алгоритмы распознавания являются основой так называемых конволюционных нейросетей [37], представляющих собой многослойные персептроны специальной архитектуры (приближенно соответствующей архитектуре зрительной коры) с конволюционными слоями, соединенными вручную, но с коэффициентами связи, найденными при обучении сети методом обратного распространения ошибок. Однако в конволюционных сетях все детекторы особенностей используют обычные (не нейросетевые) конволюционные алгоритмы распознавания.

Модели памяти, построенные на основе сверток и корреляций («неоконнекционистские» модели) описывают с единых позиций хранение в памяти индивидуальной и ассоциативной информации, а также искажения памяти. Среди них MINERVA2 [38], CHARM [39] и TODAM2 [40]. MINERVA2 хранит индивидуальные следы памяти (векторы) раздельно, но они объединяются во время процесса извлечения из памяти (в качестве механизма связывания используют операцию возведения в третью степень). CHARM является распределенной голографической моделью. Ее индивидуальные следы памяти имеют вид векторов, а их ассоциации – сверток этих векторов. Все эти свертки образуют вместе один составной след памяти; процесс вспоминания описывают перекрестные корреляции между элементами составного следа и векторами подсказок. В дополнение к индивидуальной памяти (отдельные объекты или события) и ассоциативной памяти (связывание разных представлений) TODAM2 включает в единую схему рассмотрения еще и порядковую память (память о порядке следования индивидуальных памятей). TODAM2 хранит помеченные фрагменты или «чанки» информации в виде суммы множественных *n*-кратных автосверток сумм *n* отдельных векторов. Метки и их обращения или зеркальные отображения обеспечивают доступ к чанкам и информации, которая в них хранится; индивидуальная и ассоциативная памяти являются частными случаями таких чанков с $n = 1$ и $n = 2$, соответственно. Названные модели являются биологически оправданными (с глобально функционирующими узнаванием и вспоминанием) и описывают многие явления, относящиеся к свойствам памяти, но они «не имеют реальных связей с функциями мозга» и поэтому «их реализация с помощью нейронных сетей могла бы стать следующим ярким шагом» [35, с.192].

ИНС, описанная в разд. 2, есть алгоритм свертки сама по себе и является максимально-правдоподобным алгоритмом распознавания. На примере проблемы автоматической идентификации одиночных пиков в линейчатых спектрах (компьютерная программа *PsNet*) ее детектирующие свойства были подробно изучены в [4]. В этом случае из исходного одномерного полутонового изображения («линейчатого спектра», предсталяющего собой сумму нескольких узких пиков, гладкого фона и шума) извлекают положения всех локальных его особенностей («пиков»). *PsNet* оперирует со спиноподобными бинарными данными, поэтому предварительно выполняется бинаризация исходного спектра для его преобразования в последовательность положительных и отрицательных единиц. Для *PsNet локальные рецептивные поля* имеют вид фрагментов $x_{in}$ длиною $N$ уже бинаризованного спектра; в качестве *детектора особенностей* используют ИНС длины $N$ (см. [4] и разд. 2); а *кор извлечения особенностей* представляет собой эталонный вектор $x_0$ в форме, показанной на вставке рис. 4. Чтобы найти искомые положения пиков, *PsNet* анализирует, в полном соответствии с обычной конволюционной процедурой, пересекающиеся рецептивные поля $x_{in}$. Полученная в результате анализа *карта особенностей* спектра (набор положений идентифицированных пиков) может соответствовать, например, конкретному радиоактивному ядру. Следовательно, нейросетевая программа *PsNet* [4] реализует детек тор особенностей типа свертки, выполняя процедуру извлечения особенностей аналогично тому, как это делает один из слоев конволюционной ИНС [37]. Главное отличие между ними в том, что *PsNet* в вычислительном смысле много проще и дает максимально-правдоподобные результаты распознавания (см. также разд. 14.1).

В отличие от описанных выше неоконнекционистских моделей каждая ячейка НСАМП (разд. 4), функционирует одновременно и как конволюционная, и как нейросетевая (разд. 2). Такая конволюционная (и одновременно нейросетевая) ячейка памяти хранит только один след памяти (индивидуальную память); разные индивидуальные памяти хранят в разных таких ячейках; ассоциации между ними реализованы посредством путей и связей, показанных на рис. 2 и описанных в разд. 4.2. Более того, общее число ассоциированных памятей, вообще говоря, не ограничено. Это означает, что в рамках нашей конволюционной (и одновременно нейросетевой) модели, описание индивидуальных, ассоциированных и серийных памятей осуществляется в рамках единого подхода. Обмен информацией между ассоциированными конволюционными (и одновременно нейросетевыми) памятями может быть многонаправленным (см. разд. 5) и содержание одной конкретной памяти может быть использовано как подсказка любой другой ассоциированной с нею памятью. Для нашей ансамблевой конволюционной (и одновременно нейросетевой) модели памяти в целом и большинства ее элементов существуют ясные нейробиологические аналоги (разд. 4 и 13), поэтому можно считать, что НСАМП имеет «реальные связи с функциями мозга», и реализует, пожалуй, тот «очень яркий следующий шаг» в развитии памятей типа свертки, который был предсказан в [35, с.192].

Поскольку нейросетевой и конволюционный подходы из разд. 2 как алгоритмы декодирования эквивалентны, то возникает соблазн предпринять попытку построить на их основе гипотетический вариант чисто конволюционной (без элементов нейросетей) ансамблевой модели памяти. Такая модель должна бы предполагать (см. рис. 2), что в *блоке 2* непосредственно вычисляется свертка $Q$ векторов $x_{in}$ (из *блока 1*) и $x_0$ (из эталонной памяти), после чего *блок 3* должен бы отбирать случаи $Q > 0$. Если бы это было так, то, среди прочих, немедленно возникает вопрос о специфическом нейробиологически правдоподобном механизме реализации конволюционных вычислений. Мы предпочитаем считать, что именно ИНС из разд. 2 является устройством для вычисления свертки $Q$, и поэтому упомянутый экзотический вид памяти обсуждать более не будем.

**12.4. Модульные структурно-составные нейросетевые памяти.** Общепризнано, что реальные ментальные представления в мозге являются структурными и составными (композиционными) одновременно и строятся иерархически из большого числа достаточно простых модулей согласно «правилу композиций» [41]. Это означает, что все разрешенные ментальные конструкции высокого уровня состоят из относительно малого числа составляющих низкого уровня и такая копозиционная организация является универсальным фундаментальным свойством сознания. Но среди огромного числа составных представлений высокого уровня, построенных из таких низкоуровневых составляющих простым комбинаторным способом, большинство оказываются бессодержательными и не соответствующими реальной обстановке – это так называемая «катастрофа суперпозиций» [42]. Следовательно, в мозге должен существовать некий механизм, который быстро и устойчиво обеспечивает возможность динамического связывания доступных составляющих низкого уровня в такие врéменные составные представления, которые всегда будут соответствовать постоянно меняющимся внешним условиям среды.

Нейросетевой метод является мощным средством для моделирования многих функций мозга, но, тем не менее, обычные неструктурированные ИНС, как распределенно так и плото кодированные, имеют серьезные недостатки. Во-первых, они не могут иерархически представлять и обрабатывать когнитивные символы высокого уровня, их низкоуровневые фрагменты и их взаимосвязи. Во-вторых, они не могут должным образом решать проблему обучения, в связи с чем разработчик сети вынужден либо сразу конструировать ее в соответствии с требованиями конкретной задачи (и сразу приписывать индивидуальным нейронам соответствующие им символьные значения), либо обучение системы будет требовать астрономически большого количества обучающих образов.

Обе эти задачи отчасти решены на основе представления об архитектуре динамических связей [43], использующей нейросетевые модели клеточно-ансамблевого типа (например, [44, 45, 46]). Эта идея предполагает, что скорость и фазы разрядки индивидуальных нейронов флуктуируют во времени и те наборы нейронов, которые связаны вместе для представления отдельного символа или особенности символа, образуют отдельный клеточный ансамбль. Модели связывания зрительных образов [44], выделения представляющего интерес звукового сигнала из фона [45], или продуцирования логических выводов из лингвистических предикатов [46] успешно избегают катастрофы суперпозиций путем использования упомянутого фазового механизма связывания. При этом только те нейроны (или группы нейронов) относят к целевому объекту или символу, фазы осцилляций которых связаны. Главной целью упомянутых работ было создание, избегая катастрофы суперпозиций, наиболее эффективных и наиболее приспособленных к конкретной задаче механизмов динамического связывания и динамических структурированных представлений, а не обучение синхронизованных групп нейронов, выделенных механизмом связывания. По этой причине, несмотря на довольно большой объем упомянутых сетей, их авторы конструировали их практически вручную, а потенциальная возможность упрощения процедуры обучения в структурированных сетях осталась не изученной и не реализованной. Более того, модели, основанные на методе динамического связывания, по-прежнему остаются не очень популярными, а их основная идея в большинстве работ об ИНС попросту игнорируется.

НСАМП, вводимая в настоящей работе, находится в одном ряду с динамическими модульными структурно-составными моделями, которые только что обсуждались. Однако пока мы концентрировали внимание на структуре и свойствах только элементарных модулей (модулей самого низкого уровня) нашей модульной структурно-составной модели и собственно композиционных задач касались только в иллюстративных целях (разд. 5 и 13.1). Тем не менее подчеркнем, что одношаговый алгоритм обучения, предложенный в разд. 10, практически полностью готов для его непосредственного использования в качестве исключительно адекватного и эффективного метода обучения наших ансамблевых нейросетей на любом их иерархическом уровне.

## 13. НЕЙРОБИОЛОГИЧЕСКИЕ ОСНОВЫ АНСМБЛЕВОЙ МОДЕЛИ ПАМЯТИ

При обосновании НСАМП в дополнение к синаптическому правилу Хебба (см. разд. 2 и 10), петлям обратных связей, восходящим и нисходящим нейронным путям (разд. 4, 5 и рис. 2) мы принимаем во внимание ряд перечисленных ниже нетрадиционных нейробиологических аргументов.

**13.1. Динамическая пространственно-временная синхронизация.** В последнее время получены убедительные экспериментальные доказательства того, что при выполнении испытуемыми требующих внимания когнитивных тестов небольшие группы кортикальных нейронов синхронизуют с точностью около 10 мс свою активность в области частот, относящихся к так называемой гамма-полосе ( ~ 40 Гц), — это явление *динамической пространственно-временной синхронизации*. На основе этих результатов строятся многочисленные гипотезы о том, что информация, переносимая в центральной нервной системе, скорее может содержаться в точных соотношениях между временами разрядки активных нейронов, чем в скорости такой разрядки (см. разд. 13.3), а вместилищем памяти скорее могут быть нейронные ансамбли, чем отдельные нейроны. Полагают, что один и тот же *клеточный ансамбль* — большая группа нейронов распределенная по разным областям мозга и синхронизованная в миллисекундной области времен — представляет некий объект или событие; разные клеточные ансамбли, представляющие разные объекты или события, могут сосуществовать независимо, а их возможное клеточное и функциональное перекрытие относительно мало. В разных областях и системах мозга динамическая пространственно-временная синхронизация (как для нейронной активности вызываемой стимулами, так и для текущей нейронной активности) отчасти просто отражает анатомические и функциональные связи между активными нервными клетками. Она выполняет также группирование и отбор распределенных нейронных откликов для их дальнейшего участия в обработке информации на более высоких иерархических уровнях [8, 9, 18].

Динамическая синхронизация является ключевым предположением предлагаемой модели. Когда ансамблевая память создается, консолидируется или участвует в процессе вспоминания, то механизм динамической синхронизации выделяет и активизирует в центральной нервной системе распределенный клеточный ансамбль, соответствующий конкретному стимулу. Этот стимул (объект, событие или правило) представляется в виде *набора* ассоциированных ячеек ансамблевой памяти, каждая из которых соответствует отдельной особенности стимула, а все вместе они соответствуют ансамблевой памяти о стимуле *в целом* (разд. 4 и 5). Когда клеточный ансамбль, относящийся к стимулу представляющему текущий интерес, выделен (синхронизован), то он автоматически отбирает соответствующий ему синхронизованный набор входных сигналов, используя для этого временное окно, ширина которого соответствует точности синхронизации (разд. 1 и 4.4). В результате число спайков, которые могут оказывать влияние на конкретную ячейку ансамблевой памяти, резко уменьшается, а итоговый набор из $N$ синхронизованных входных импульсов является по сути *сообщением* в форме $N$-мерного бинарного вектора, которое адресовано конкретной ячейке ансамблевой памяти (см. разд. 1).

**13.2. Размер основной сигнальной ячейки.** Путем анализа отношения сигнал/шум внутри кортикальных колонок было оценено число синаптических контактов кортикальных нейронов – это 3000-10000, в то время как размер основной сигнальной ячейки в церебральной коре составляет, согласно этим же оценкам, всего около 100 нейронов [47]. Если предположить, что нейроны сети связаны между собой по правилу «все со всеми», то число синаптических контактов на один нейрон можно интерпретировать как полное число $N_{sps}$ нейронов (входных спайков), которые могут участвовать в процессе отбора импульсов, попадающих во временное окно синхронизованного клеточного ансамбля (*блок 1* на рис. 2), а размер сигнальной ячейки может соответствовать числу спайков $N_{dns}$, отобранных как результат такой синхронизации, или, иначе, $N = N_{dns}$ можно рассматривать как размерность входных/выходных векторов клеточного ансамбля.

С другой стороны, на один кортикальный нейрон проецируются только 10-100 нейронов таламуса (все сенсоры, за исключением обонятельного, передают информацию к церебральной коре через таламус). В этом причина, например, того, что один нейрон из слоя 4 в первичной зрительной коре кошки получает входы только от приблизительно 30 из возможных 360000 латеральных геникулярных ядер (нейронов, которые проецируются в церебральную кору). В среднем же процент таламических спайков, вызывающих один кортикальный спайк, и процент кортикальных спайков, запускаемых таламическими спайками, оценивают в ~1-10% [48]. Из эксперимента также известно, что ансамбль из всего 12-16 передаточных нейронов зрительной системы кошки достаточен для удовлетворительной реконструкции находящихся в поле ее зрения реальных движущихся сцен [49], а активность всего 50-100 кортикальных моторных нейронов может управлять одно- и трехмерными движениями рук робота с очень хорошим качеством воспроизведения исходных движений живого организма [50, 51]. Достаточно хорошее качество предсказания движений руки робота дает и гибридный интерфейс мозг-машина, использующий управ ляющие сигналы всего от ~18 нейронов [52]. Таким образом, разные результаты современной нейрофизиологии согласуются с предположением о том, что для конкретной ансамблевой ячейки памяти (рис. 2) число *N* ее входных/выходных нейронов может быть, наиболее вероятно, около 100.

**13.3. Раннее прецизионное испускание спайков и их распределенные всплески.** Нейронные коды, используемые живыми организмами для кодирования стимулов и извлечения их особенностей, исследуют путем фиксации моментов испускания спайков как отдельными нейронами, так и их относительно небольшими популяциями. Например, при изучении зависящих от времени стимулов найдено, что зарегистрированные моменты испускания всех спайков от одного активного нейрона, несут большее количество информации, чем моменты испускания только изолированных спайков (спайков между их всплесками), и качество извлечения особенностей может быть дополнительно улучшено, если учитывать только те спайки, которые возникают во время самих всплесков (компактных групп спайков) [34]. Но наилучшие результаты при извлечении особенностей зависящих от времени стимулов были получены при анализе совпадающих спайков от пар активных нейронов [53]. Совокупность этих эмпирических фактов стала основой гипотезы о том, что коррелированная во времени спайковая активность группы нейронов может рассматриваться как ее *распределенный всплеск* [53], который несет наиболее полную информацию об особенностях меняющихся стимулов и, в свою очередь, может быть сформирован с помощью так называемого «наилучшего нейронного кода» [11, 53].

Если *популяция* кортикальных нейронов активируется синхронизованными последовательностями токовых импульсов, распределенных почти по Пуассону и характеризующихся одинаковой средней частотой, то результирующий набор нейронных откликов проявляет высокую вариабельность по частоте и содержит обособленные всплески спайков. В получаемом *наборе* последовательностей нервных импульсов (соответствующем числу нейронов популяции) первые спайки большинства всплесков оказываются выстроенными – их временной разброс составляет 1-5 мс и уменьшается при увеличении степени исходной синхронизации. Однако в период задержанной нейронной активности (через ~20 мс после начала всплеска) моменты испускания спайков от разных нейронов популяции становятся практически случайными. Это явление *раннего прецизионного испускания спайков* показывает, что информацию, содержащуюся в группе синхронизованных нейронов, могут переносить или *первые прецизионно синхронизованные* спайки нейронных откликов, или *средняя скорость* испускания спайков, если рассматривать только позднюю (задержанную) активность группы нейронов [54]. В других опытах дополнительно было установлено, что как для отдельных нейронов, так и для пар коррелированных нейронов момент регистрации только *первого* спайка, появившегося после предъявления стимула, может нести до 90% информации, которая содержится во всей цепочке инициированных стимулом нервных импульсов [12].

Результаты компьютерного моделирования сетей нейронов с близкими к реальным физиологическими характеристиками поддерживают выводы таких экспериментов. Например, компьютерное моделирование активности сетей из случайно связанных нейронов с зависящими от частоты свойствами синапсов показывает, что спайковая активность таких сетей может *самоорганизоваться* как спонтанно, так и под влиянием внешнего стимула, что приводит к образованию синхронизованного пакета спайков, включающего отклики от 95-98% нейронов сети. Это так называемый *популяционный всплеск* – большая группа нейронов с активностью, синхронизованной в узком временном окне (~5 мс). Причем такие спонтанные или связанные с предъявлением стимула популяционные всплески нейронной активности, перемежающиеся периодами асинхронного испускания спайков, возникают даже при рассмотрении малых сетей без предположения о наличии в них какой-либо специфической структуры. Обязательным условием является только наличие межнейронных взаимодействией с нелинейными (близкими к реальным) синапсами [10].

В пределе малых времен наблюдения задача о переносе информации в нейронных сетях исследовалась и в рамках строгого модельно-независимого подхода, основанного на шенноновской теории информации. При этом, в частности, было найдено, что вся информация, переносимая популяцией *невзаимодействующих* нейронов, содержится только в *средней скорости* разрядки отдельных нейронов, а для *коррелированных* нейронных популяций вся переносимая ими информация содержится только в их *корреляциях* и вовсе не содержится в средней скорости разрядки нейронов [55].

Ясно, что *первые прецизионно выстроенные спайки* всплесков активности группы синхронизованных нейронов [12, 54], *популяционные* [10] или *распределенные всплески* [53] такой активности являются нейробиологическими аналогами характеристических векторов $x(d)$ и совместно образуют для этих векторов (см. разд. 1, 4.1 и др.) серьезную нейробиологическую основу.

**13.4. Устойчивое распространение спайковой активности.** Результаты компьютерного моделирования, выполненного с использованием модельных нейронов с реальными физиологическими и анатомическими параметрами, подтверждают также возможность стабильного распространения синхронизованной спайковой активности в сложных кортикальных нейронных сетях, если набор таких спайков достаточно велик (около 100) [56]. Из этих же расчетов следует, что нейроны, участвующие в переносе синхронизованной спайковой активности, на разных этапах этого процесса могут быть разные и с этим фактом хорошо согласуется предполагаемый моделью механизм динамического формирования путей для распространения характеристических векторов $x(d)$, описанный в разд. 4.2. Таким образом, результаты компьютерного моделирования в нейрофизиологических условиях, близких к реальности, поддерживают предположение о том, что как внутри, так и вне конкретной ячейки ансамблевой памяти (рис. 2) перенос информации группой синхронизованных спайков (ее модельный аналог – вектор $x(d)$) возможен, что эта группа может распространяться устойчиво, и что она может состоять приблизительно из 100 импульсов (т.е. размерность $N$ пространства векторов $x(d)$ может быть порядка 100).

**13.5. Нейроны зависящие от времени и нейроны-детекторы ошибок.** Для зависящего от времени модельного описания процесса извлечения информации из памяти важно знать естественный и биологически оправданный способ измерения времени, начиная с момента начала процесса вспоминания. Следуя работе [57], мы заимствуем этот способ из результатов современных нейрофизиологических исследований, полученных при исследовании кратковременной памяти обезьян путем регистрации скорости испускания спайков их кортикальными нейронами (см., например, [58, 59]). В таких экспериментах были обнаружены *нейроны зависящие от времени*: их спайковая активность скачкообразно возрастает в момент появления кратковременного стимула (это «инициирующее событие»), после чего средняя скорость их разрядки линейно уменьшается в течение *характерного времени* или *периода распада* $t_0$, которое может меняться в пределах от десятков миллисекунд до десятков секунд. Именно $t_0$ мы используем как естественную временную шкалу для описания зависящих от времени процессов памяти (см. разд. 4 и рис. 2).

Используя для регистрации нейронной активности интрацеребральные электроды, постоянно внедренные в глубокие структуры человеческого мозга, были выполнены уникальные эксперименты, в ходе которых испытуемые выполняли некоторые когнитивные задания, связанные с языком и памятью (см., например, [60]). Таким способом в некоторых подкорковых областях мозга (субкортикальные ядра таламуса, стриопаллидальная система и т.д.) были обнаружены «*нейроны-детекторы ошибок*» – нейронные популяции, которые изменяют среднюю скорость своей разрядки избирательно, только когда сделана ошибка при выполнении испытуемым когнитивного теста. В нашей модели памяти они используются, в частности, при создании блока сравнения петли экспликативной обратной связи (разд. 4.3 и *блок 5* на рис. 2).

## 14. НЕКОТОРЫЕ ПРИМЕНЕНИЯ АНСАМБЛЕВОЙ МОДЕЛИ ПАМЯТИ

Рассмотрим применения НСАМП к решению ряда междисциплинарных задач, для которых конкретные количественные и качественные результаты можно получить без проведения дополнительных вычислений и которые не связаны с архитектурой компьютеров или другими чисто техническими аспектами (см. также разд. 11.5).

**14.1. Максимально-правдоподобное распознавание образов и теория зрения.** Как видно из вставки на рис. 4, эталонный вектор $x_0$ можно интерпретировать как дискретное представление короткого «белого отрезка» на «черном фоне» (компоненты −1 «вокруг» компоонент +1 вектора $x_0$). Следовательно, функции $P(d)$ на рис. 4 можно рассматривать как результат численного исследования различных вариантов нейросетевого алгоритма из разд. 2 для максимально-правдоподобной идентификации названных одномерных бинарных (спиноподобных) образов, искаженных бинарным шумом. Такая задача возникает, например, после бинаризации линейчатых спектров излучения (или других полутоновых изображений) при идентификации пиков нейросетевой программой *PsNet*, свойства которой как алгоритма распознавания были подробно исследованы в [4].

Качество поиска пиков компьютерной программой или человеком описывает условная вероятность их истинного обнаружения $D(A/Bkgr)$, где $A$ – площадь пика, а $Bkgr$ – среднее значение фона под ним, при условии, что условная вероятность $\alpha$ ложного обнаружения пиков постоянна [3, 4]. Поскольку $A/Bkgr$ можно интерпретировать как отношение сигнал/шум, то $D(A/Bkgr) = P(q)$ (во всяком случае пока полная ширина пика у его основания меньше $N$) и $P(q) = \alpha$, если $q = 0$ (см. разд. 7) Для опытных операторов и дилетантов ($\alpha = 0.012 \pm 0.004$ и $\alpha \sim 0.3$, соответственно [3]) измеренные функции $P(q)$ – вероятности истинной идентификации пиков [3] или *психометрические функции* [32] – количественно воспроизводятся программой *PsNet* полностью (см. рис. 6 работы [4]). Похожие психометрические функции изучают уже 100 лет [32], но, насколько нам известно, наши результаты – это первый и пока единственный пример, когда конкретные данные экспериментального тестирования зрительной системы человека теоретически описываются полностью. Прекрасное количественное согласие между результатами измерений и вычислений позволяет предположить, что нейросетевой максимально-правдоподобный алгоритм распознавания из разд. 2 действительно может использоваться зрительной системой человека для распознавания те же образов, что и соответствующая компьютерная программа.

Предложенный подход к максимально-правдоподобному решению одномерных задач распознавания можно легко обобщить на случай распознавания объектов произвольной топологии и размерности (например, распознавание букв или цифр).

**14.2. Нейропсихология.** Из нейропсихологических экспериментов известно, что фронтальные пациенты (пациенты с *локальными* повреждениями фронтальных долей мозга) практически полностью сохраняют способность к узнаванию, а их способность к вспоминанию существенно ослаблена. Как видно из рис. 4, для неповрежденных ИНС (кривая 1) их вероятности вспоминания $P(d)$, $0 < d \leq 1$, в целом больше, чем для сетей с локальными повреждениями (кривые 2-6), но для всех названных сетей значение вероятности узнавания одинаково, $P(0) = 1$. Поскольку в рамках НСАМП память фронтальных пациентов моделируется сетями с локальными повреждениями, то, следовательно, модель предсказывает сохранение способности фронтальных пациентов к узнаванию и ослабление их способности к свободному вспоминанию и вспоминанию с подсказкой. Этот результат полностью согласуется с экспериментальными данными о свойствах *эпизодической памяти* человека [33].

Согласно НСАМП свободное вспоминание, вспоминание с подсказкой и узнавание являются частными случаями единого универсального механизма извлечения информации из памяти (разд. 4 и 7, рис. 2 и 4). Действительно, вспоминание − это *стохастический* процесс, инициируемый *серией* случайных или отчасти случайных векторов $x_{in}$ поступающих на вход обученной ИНС, в ходе которого каждый очередной результат извлечения из памяти $x_{out}$ сравнивается с эталоном $x_0$. Серия разных случайных векторов $x_{in} = x_r$ инициирует *свободное вспоминание* ($q = 0$), серия отчасти случайных векторов $x_{in} = x(1 − q)$ инициирует *вспоминание с подсказкой* ($0 < q < 1$), детерминированный вектор $x_0$ инициирует *узнавание* ($q = 1$), но во всех этих случаях используется одна и та же ячейка ансамблевой памяти. Таким образом, узнавание и все виды вспоминания отличаются только *информацией*, инициирующей извлечение из памяти или *интенсивностью* используемой подсказки. При этом системы (области) мозга, вовлеченные в процесс, в целом, одни и те же. Следовательно, наша модель в своем простейшем варианте поддерживает предположение о существовании *тесной взаимосвязи* между вспоминанием и узнаванием [61] и не вполне согласуется с представлением о том, что они зависят от *разных* систем мозга [62].

**14.3. Психолингвистика.** *Словарный частотный эффект* (СЧЭ) исследуют экспериментально и теоретически как при воспроизведении речи, так и при ее восприятии. Экспериментально установлено, что разница в *задержках воспроизведения* редких и часто встречающихся слов легко достигает значений $T = 50$-$100$ мс. Основной вывод теоретических исследований состоит в том, что СЧЭ возникает на уровне доступа к *словарным формам* [63].

НСАМП предполагает *зависящий от времени* механизм извлечения из памяти и поэтому дает шанс описать СЧЭ количественно. Полагаем, что нейросетевая память (*блок 2* на рис. 2), относящаяся к конкретной ячейке ансамблевой памяти, хранит словарную форму конкретного слова, а функция $P(d)$ описывает характеристики этой нейросетевой памяти. Соотношение между свойствами памяти о словарных формах для часто встречающихся слов, $P_{fr}(d)$, и редко встречающихся слов, $P_{if}(d)$, должно быть $P_{fr}(d) \geq P_{if}(d)$, если $0 < d \leq 1$ (вспоминание), или $P_{fr}(d) = P_{if}(d)$, если $d = 0$ (узнавание). Например, на рис. 4 кривые 1 и 6 могут соответствовать $P_{fr}(d)$ и $P_{if}(d)$, соответственно. При условии, что интенсивность подсказки $q = 1 − d$ постоянна, число векторов $x_{in}$, необходимых для извлечения из памяти словарных форм часто и редко встречающихся слов, обозначим $n_{fr}$ и $n_{if}$, соответственно. Тогда задержка при их воспроизведении будет объясняться просто разным количеством векторов $x_{in}$, необходимым для извлечения из памяти соответствующих словарных форм: $\Delta n = n_{if} − n_{fr}$. Принимая во внимание это предположение, значение $T$ и типичный диапазон частот эффекта динамической пространственно-временной синхронизации, $f = 20$-$90$ Гц, число $\Delta n$ легко оценить: $\Delta n = fT$. Простые вычисления приводят к целочисленной оценке $0 < \Delta n < 10$.

Только что вкратце изложенная ансамблевая нейросетевая модель СЧЭ дает единое описание воспринимаемых слов (восходящий поток входных сигналов) и воспроизводимых слов (нисходящий поток входных сигналов) и поэтому выглядит как реакция на одно из предложений для будущих исследований из работы [63].

Наша модель стала также основой количественной нейросетевой модели [64] явления на кончике языка (НКЯ), когда человек временно не может вспомнить хорошо известное ему слово (в художественной форме яркий случай проявления такого затяжного состояния описан в известном рассказе А.П.Чехова «Лошадиная фамилия» [65]). Согласно НСАМП конкретному состоянию НКЯ соответствует кривая 4 на рис. 4 (с пологим плато), так как она описывает зависимость вероятности $P$ вспоминания слова от интенсивности подсказки $q$, типичную для человека в состоянии НКЯ: относительно малые подсказки мало помогают вспоминанию (пологое плато на кривой 4) и только при больших подсказках, когда $q \approx 1$, $P(q)$ быстро растет до значения $P = 1$ (как было показано в разд. 4.2 фактически это означает узнавание). Предложенная количественная модель предоставляет возможность определять и вычислять как силу переживаемых состояний НКЯ так и вероятность их появления, объединить психолингвистический подход к анализу НКЯ с подходами, основанными на анализе памяти и метапамяти, перебросить мостик между традициями исследования НКЯ на основе речевых ошибок и хронометрии наименования слов (см. выше), и, наконец, количественно описать [66, 67] поведение человека в таких сложных ситуациях, как, например та, которую переживает персонаж упомянутого рассказа А.П.Чехова.

**14.4. Психология эмоций и чувств.** Ансамблевая модель памяти допускает раздельное описание сознательных и подсознательных ментальных процессов способом, явно зависящим от времени (см. разд. 4.6 и рис. 2). Используя этот факт и имеющиеся данные о нейрофизиологии и эволюционной биологии эмоций и чувств [17], нами предложена новая концептуальная модель проявления эмоций и чувств и их первая количественная нейросетевая модель, которая была применена для количественного описания [68] *чувства знания* (ЧЗ) или *чувства происходящего*, если пользоваться терминологией работы [17]. Согласно модели, возникновение ЧЗ происходит на этапе активизации относящейся к вспоминаемому слову ансамблевой ячейки памяти в момент идентификации в метапамяти (ее роль играет эталонная память) эталонного следа вспоминаемого слова. Когда речь плавная, то ЧЗ не успевает осознаваться, так как каждое очередное вспоминаемое слово извлекается из памяти при функционировании только *импликативной* петли обратной связи 1-2-3-4-1. Зато в состоянии НКЯ человек вполне осознает возникшее у него ЧЗ. Это происходит, когда неудачей заканчивается первый полный цикл импликативного вспоминания, и в *блоке 5* проверяется, существуют ли *экспликативная* (осознаваемая и связанная с окружением) причина для продолжения имликативного вспоминания. Вообще говоря, модель [68] предоставляет возможность для объяснения любых явления, связанных с эмоциями, чувствами или настроениями и у живутных, и у человека, но главный ее вывод состоит в том, что эмоции не разделяют осознаваемые и неосознаваемые ментальные процессы, а только создают для них соответствующий эмоциональный фон.

## ЗАКЛЮЧЕНИЕ

Показано, что триарно-бинарный алгоритм кодирования-декодирования [1] обладает свойствами максимального правдоподобия и на его основе предложена новая нейросетевая ансамблевая модель памяти. Ее главными отличительными чертами являются максимально-правдоподобные основные характеристики памяти (вероятности вспоминания-узнавания) и новая архитектура ячейки памяти, включающая двухслойную хопфилдовскую сеть, *N*-канальные временные ворота, дополнительную эталонную память и две вложенные петли обратной связи. В целом предлагаемая модель является структурно-составной и строится иерархически, но в центре внимания настоящей работы были только свойства ее модуля самого низкого уровня – нейросетевой ансамблевой ячейки памяти.

Для лежащего в основе модели метода кодирования найден такой вариант хопфилдовской ИНС, который реализует максимально-правдоподобный алгоритм декодирования типа свертки и, одновременно, линейный классификатор спиноподобных бинарных векторов по отношению к их хемминговскому расстоянию до вектора-эталона. Другими словами найдены условия, при которых нейросетевой, конволюционный и хемминговский подходы к анализу бинарных данных эквивалентны и реализуют один и тот же максимально-правдоподобный алгоритм декодирования, вспоминания, узнавания, обнаружения или классификации. Найден эффективный метод одношагового обучения используемых сетей, метод точного (или приближенного, но с любой наперед заданной точностью) чис ленного определения модельных вероятностей вспоминания-узнавания и, если сеть не повреждена, точные аналитические формулы для таких вычислений.

Источником всех достоинств предложенной модели памяти являются достоинства используемого алгоритма кодирования-декодирования [1]. Мы предполагаем, что этот алгоритм может реализовывать тот «наилучший нейронный код» и быть основой того «идеального» нейронного кодера-декодера, которые сейчас активно ищут с целью установления природы вычислений в нервных тканях животных и человека. Наряду с результатами нашей модели это предположение поддерживают выводы нейрофизиологических экспериментов, результаты теории информации и компьютерного моделирования сетей с близкими к реальным нейрофизиологическими параметрами.

В дополнению к основным характеристикам памяти, активным и пассивным следам памяти и способам их взаимных преобразованииий наша нейросетевая модель позволяет описать ослабления памяти разной природы, активизацию памяти и ее одношаговое обучение, явную зависимость от времени процесса извлечения информации из памяти, а также предоставляет возможность моделирования метапамяти, обобщенного (не выражаемого словами) представления знаний и раздельного описания осознаваемых и неосознаваемых ментальных процессов. Показано, что ячейка нейросетевой ансамблевой памяти может рассматриваться как наименьший неделимый элемент сознания и использоваться как кирпичик для построения более сложных форм осознаваемой памяти или осознаваемых когнитивных функций мозга в целом. Модель явно предполагает, что процесс активизации памяти предшествует ее работе в режиме обучения или вспоминания, а получаемое в результате активизации состояние активной памяти можно сопоставить введенной Е.Тулвингом «моде извлечения из памяти» [28]. Предложенная модель памяти стала основой количественных компьютерных моделей некоторых явлений из различных областей знания, среди которых распознавание образов и теория зрения, нейропсихология, психолингвистика и психология эмоций и чувств. Мы рассматриваем этот факт как весомый аргумент в поддержку ее правильности.

В рамках модели пассивные следы памяти (информация, долговременно хранимая в ячейке памяти) и активные следы памяти (та же информация, переносимая с целью ее обработки) имеют одинаковую (бинарную) цифровую форму. Следовательно, каждая ячейка ансамблевой памяти может одинаково успешно хранить следы памяти любой природы, например, об объектах, событиях, символах или правилах. Таким образом, модель явно предполагает, что обработка информации в нервных тканях животных и человека выполняется в бинарной арифметике и это факт может стать основой для естественного преодоления известного противоречия [69] между локалистским (символьным) и коннекционистским (классическим нейросетевым) подходами к моделированию памяти и других когнитивных функций мозга.

Для нейробиологического обоснования НСАМП в дополнение к хеббовскому синаптическому правилу, петлям обратной связи, восходящим и нисходящим потокам нервных импульсов в мозге использованы следующие нетрадиционные нейробиологические аргументы: явление динамической пространственно-временной синхронизации, размер основной сигнальной ячейки нейронов, раннее прецизионно синхронизованное испускание спайков группой нейронов, популяционные или распределенные всплески спайковой активности, устойчивое распространение в кортикальных сетях синхронизованной спайковой активности нейронов, свойства зависящих от времени нейронов и нейронов-детекторов ошибок. Отдельные элементы модели и модель в целом имеют ясные аналогии в биологии и компьютерной технике, хотя эталонная память является тем новым элементом модели, вероятная нейробиологическая природа которого еще должна быть установлена в будущем. Более того, модель может решать и некоторые так называемые «тяжелые» проблемы моделирования памяти (например, дилемму стабильности и пластичности или образцов и прототипов [70]) простым и радикальным способом: они попросту отсутствуют в рамках предлагаемого подхода. Но эта и многие другие грани модели являются предметом для последующего обсуждения.

# A NEURAL NETWORK ASSEMBLY MEMORY MODEL WITH MAXIMUM-LIKELIHOOD RECALL AND RECOGNITION PROPERTIES

*P.M.Gopych*

*Kharkiv National University, Kharkiv 61077, Ukraine, pmg@kharkov.com*